\documentclass[12pt]{report}
\usepackage[numbers,sort&compress]{natbib}
\usepackage{geometry}
\usepackage{fancyhdr}
\usepackage{afterpage}
\usepackage{graphicx}
\usepackage{amsmath}
\usepackage{amssymb,amsbsy}
\usepackage{dcolumn,array}
\usepackage{tocloft}
\usepackage{asudis}
\usepackage{acronym}
\usepackage{multirow}
\usepackage[space]{grffile}
\usepackage{xcolor}
\usepackage{morefloats}
\usepackage{footnote}
\usepackage[ruled,vlined]{algorithm2e}
\usepackage{cleveref}
\usepackage{textcomp}
\usepackage{url}
\usepackage{mathtools}

\DeclareMathOperator*{\argmin}{argmin}

\newcommand\myeq{\stackrel{\mathclap{\normalfont\mbox{{\tiny i.i.d}}}}{\sim}}
\graphicspath{{./Figures/}}

\usepackage{enumitem}
\usepackage{caption}
\usepackage{subcaption}
\usepackage[utf8]{inputenc}
\usepackage{appendix}
\usepackage{breakurl}

\makeatletter
  \@addtoreset{chapter}{part}
  \@addtoreset{@ppsaveapp}{part}
\makeatother
\newtheorem{theorem}{Theorem}
\newtheorem{defn}{Definition}
\allowdisplaybreaks

\let\ref\Cref
\begin{document}
\pagenumbering{roman}
\title{Bayesian Nonparametric Reinforcement Learning \\ \ \\ in LTE and Wi-Fi Coexistence}
\author{Po-Kan Shih}
\degreeName{Master of Science}
\defensemonth{April}
\gradmonth{May}
\gradyear{2021}
\chair{Bahman Moraffah}
\cochair{Antonia Papandreou-Suppappola}
\memberOne{Gautam Dasarathy}
\memberTwo{YiChang Shih}
\maketitle
\doublespace
\begin{abstract}
With the formation of next generation wireless communication, a growing number of new applications like internet of things, autonomous car, and drone is crowding the unlicensed spectrum. Licensed network such as the long-term evolution (LTE) also comes to the unlicensed spectrum for better providing high-capacity contents with low cost. However, LTE was not designed for sharing spectrum with others. A cooperation center for these networks is costly because they possess heterogeneous properties and everyone can enter and leave the spectrum unrestrictedly, so the design will be challenging. Since it is infeasible to incorporate potentially infinite scenarios with one unified design, an alternative solution is to let each network learn its own coexistence policy. Previous solutions only work on fixed scenarios. In this work a reinforcement learning algorithm is presented to cope with the coexistence between Wi-Fi and LTE agents in 5 GHz unlicensed spectrum. The coexistence problem was modeled as a decentralized partially observable Markov decision process (Dec-POMDP) and Bayesian approach was adopted for policy learning with nonparametric prior to accommodate the uncertainty of policy for different agents. A fairness measure was introduced in the reward function to encourage fair sharing between agents. The reinforcement learning was turned into an optimization problem by transforming the value function as likelihood and variational inference for posterior approximation. Simulation results demonstrate that this algorithm can reach high value with compact policy representations, and stay computationally efficient when applying to agent set.
\end{abstract}
\begin{acknowledgements}
There are a number of people to whom I would like to dedicate my uttermost appreciation in the journey of pursuing my degree at Arizona State University. First and foremost, it is my advisors, Bahman and Antonia. To Bahman, you are more than an advisor, but a mentor and a friend. Your supervision has inspired my interest and has been pushing forward my research. Thank you for advising not only on academic work, but also on life and future careers. Thank you for opening the doors of Bayesian inference and reinforcement learning for me, as well as training me to be a problem solver. Your encouragement always reminds me never to flinch every time I suffer from frustration. To Antonia, thank you for being teaching me random signal theory and signal processing, which established my background for my research, and inviting me to your house. Your kindness and humor always relax me when I feel nervous. To Gautam, thank you for teaching me theory behind machine learning, and being my committee members. To YiChang, thank you for your time and advice in my defense.

I also need to thank Si-Hua, one of my best friend. Although we could not meet in person, you still keep in touch with me through network. Our communication makes me feel close to my friends in Taiwan. Your messages always warm my heart. In no uncertain terms, I am indebted to my family in Taiwan. Since the first day I came here, mother, father, and sister, you have been my strongest backing. Thank you for giving me a sound mind, keeping reminding me to take care of myself even if we are tens of thousands of miles apart. Thank you for being my best listeners whenever I feel down. Our video calls have been alleviated my loneliness, especially in these tough days. To friends from Arizona State University - Henry, Shawn, Richard, and Russell - thanks for your help in projects and coursework and many funny moments during these years. To everyone who helped me, thank you for making me feel at home.
\end{acknowledgements}
\tableofcontents
\addtocontents{toc}{~\hfill Page\par}

\addcontentsline{toc}{part}{LIST OF TABLES}
\renewcommand{\cftlabel}{Table}
\addtocontents{lot}{Table~\hfill Page \par}
\listoftables
\newpage
\addcontentsline{toc}{part}{LIST OF FIGURES}
\renewcommand{\cftlabel}{Figure}
\addtocontents{lof}{Figure~\hfill Page \par}

\addtocontents{toc}{CHAPTER \par}
\listoffigures


\doublespace
\pagenumbering{arabic}
\chapter{INTRODUCTION}
\label{chap:intro}
\begin{figure}
\centerline{\includegraphics[width=1\linewidth]{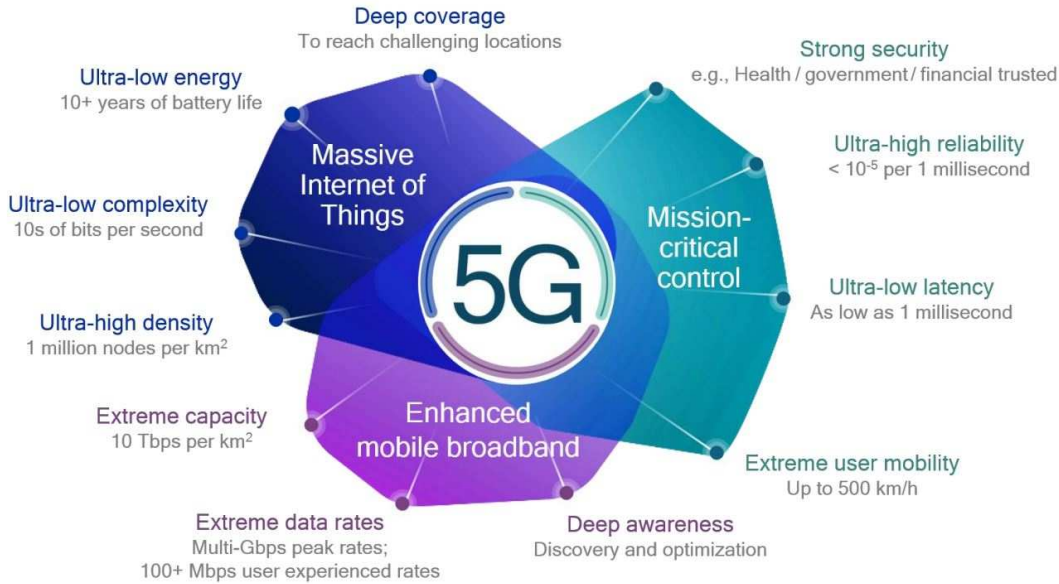}}
	\caption[The Vision of 5G Network]{The 5G wireless network is anticipated to provide various high-quality services\footnotemark.}
	\label{fig:5G_vision}
\end{figure}

\acresetall
With the population of wireless devices growing exponentially comes the massive demand for spectrum resources in the fifth generation wireless network (5G). As \ref{fig:5G_vision} illustrates, one ambition of the 5G network is to fulfill the requirements for various ultra-dense, scalable, and highly customizable networks while boosting the throughput. To satisfy this, it is essential for the cellular networks to provide more capacity but not to raise the operational costs significantly. Since the licensed spectrum is limited and has been crowded, the unlicensed spectrum is attracting attentions from network operators. Offloading to the unlicensed spectrum provides two major advantages: access flexibility for unexpected incoming loads and cost efficiency since it is free. According to the Cisco annual internet report \cite{cisco}, the number of Wi-Fi hotspots is expected to be up to 628 million, and the number of cellular network subscribers will reach 5.7 billion by 2023. However, currently many heterogeneous wireless networks have been crowding the unlicensed spectrum, including Wi-Fi, bluetooth, various internet of things (IoTs), and other new applications, such as autonomous car, radar, drone, still keep arriving. These applications suggest potentially infinite number of wireless devices are entering and leaving the spectrum continually. The existing networks are divergent in their properties such as quality of services (QoS), protocols, bandwidth requirements, and access timing. The nature that resource requests are not constant obstructs the coordination between networks. For example, wireless wide area networks such as the Long-Term Evolution (LTE) demand stable and massive throughput for its high-quality, heavy-load content, whereas countless cloud-integrated IoTs devices may require frequent and burst-like service. This problem has motivated the study for spectrum sharing techniques \cite{bg_5Gunlincesed,bg_HAN201653,bg_wifiLTEDC}.
%

In this work, we consider a fair spectrum sharing between Wi-Fi and LTE networks in the 5 GHz unlicensed band. An infinite-horizon decentralized partially observable Markov decision process (Dec-POMDP) is adopted to simulate the interaction between wireless nodes and the spectrum environment. A cumulative reward function is proposed to measure the quality of sequential decision on the competition for limited spectrum resource. We utilize an off-policy, mode-free reinforcement learning to learn policies for each agent from episodes collected by behavior policy. To accommodate the various policy representations for different types of wireless nodes, our reinforcement learning applies Bayesian approach to infer the distribution over policies with nonparametric policy prior. For posterior model approximation, the variational inference is utilized in consideration of model complexity. To our best knowledge, this is the first work on spectrum sharing for LTE and Wi-Fi utilizing reinforcement learning with variable policy representations.
\footnotetext{From \url{https://www.qualcomm.com/media/documents/files/whitepaper-making-5g-nr-a-reality.pdf}.}

The rest of this work is organized as follows. In chapter 2 we first review previous researches on spectrum sharing and applications of Bayesian nonparametric models on signal processing. Then the nonparametric model utilized in this work is introduced. Markov chain Monte-Carlo method and variational inference are two major approaches to estimate the posterior model in Bayesian inference; their fundamentals and implementations, the Gibbs sampling and coordinate ascent variational inference, will be presented in this chapter. Reinforcement learning is an optimization process based on the (partially-observable) Markov decision process model, which components will be discussed in the last section of chapter 2. Chapter 3 exhibits our approach of modelling a spectrum sharing problem to partially-observable Markov decision process, and elaborates the iterative algorithms of proposed Bayesian inference for policy learning. Performance evaluation of proposed algorithm in comparison with previous method is demonstrated with discussion in chapter 4. Chapter 5 summarizes our contributions and proposes some future extension of this work.
\chapter{BACKGROUND}
\label{chap:background}
\acresetall
In this chapter, we are going to discuss background of the problem and some underlying techniques in our method. First, review on recent researches about application of reinforcement learning on spectrum sharing is delivered. Then the fundamentals of Bayesian nonparametric model utilized and its application in signal processing are introduced. For discrete cases, Dirichlet process is a classical model widely-used in Bayesian inference. Dirichlet process has several convenient realization methods, including the Chinese restaurant process and the stick-breaking process.. In order to gather information from the posterior distribution, Markov chain Monte Carlo method is a direct way to draw samples from it, and Gibbs sampling is one implementation of this theory. Although sampling method like Markov chain Monte Carlo can deliver accurate statistical information about the distributions, its resource-demanding property indeed restricts its application to problems with high dimension. Variational inference, on the other hand, provides a less accurate but faster alternative to approximate the posterior distributions. In the last section the Markov decision process, the cornerstone of reinforcement learning, and how reinforcement learning optimizes it, will be introduced.
\section{Spectrum Sharing}
\label{sec: SS_litreview}
Spectrum sharing has been a popular research topic. Some mechanisms have been employed \cite{LTEWiFisurvey} in existing networks. The Carrier Sense Multiple Access/Collision Avoidance (CSMA/CA) was encoded in IEEE 802.11 Wi-Fi standard to handle the homogeneous coexistence for Wi-Fi access points and user equipments. CSMA/CA is an uncoordinated scheme, which incorporates Lister-Before-Talk (LBT) mechanism to avoid collisions. LBT requires each node (could be access point or user equipment) to perform spectrum sensing for the channel it is going to access before transmission commences. By sensing before transmission, new coming node suspends its transmission when the channel is sensed busy. Even if the channel is sensed idle, the node is not allowed to access the channel immediately, since there might be other nodes waiting to utilize the same channel. All waiting nodes start their transmission immediately after channel becomes idle will cause severe collision. To avoid coincident transmission, CSMA/CA mandates node to execute back-off sensing for several time slots, and transmission can start only when the back-off sensing result is idle for the given time slots. With the number of time slots being stochastic for each node, probability of colliding transmission is reduced. As we mentioned in \ref{chap:intro}, with the increasing demand for high-quality, low-latency contents, LTE operators seek to expand their spectrum usage to unlicensed spectrum. Some LTE-unlicensed (LTE-U) mechanisms have been proposed for LTE networks to operate in unlicensed spectrum. LBT and Almost Blank Subframe (ABS) are two branches. LBT-based mechanism can be deployed in areas like Europe and Japan where channel assessment before transmission is mandatory, whereas ABS-based mechanism can be implemented immediately in areas without requirement of sensing before transmission, such as United States, China, Korea, and India. Unlike aforementioned LBT, ABS employs duty-cycle for LTE nodes to share spectrum with other networks. LTE node actively interrupts its transmission for other networks to access the channel in every period of time, and the interrupt period depends on the measurement of channel occupancy. Compare to LBT mechanism, ABS-based method needs less channel sensing actions and less modification on current LTE framework in exchange of higher probability of collision. For standardization, ABS-based mechanism has been incorporated in 3GPP LTE release 10. Among all candidate methods, the LTE-Licensed Assisted Access (LTE-LAA) is one of the most competitive schemes because its operation is similar to Wi-Fi and can be adopted in all regions in the world. The LTE-LAA adopts LBT mechanism, and has been standardized in 3GPP LTE release 13 to offload downlink traffic for LTE to the unlicensed spectrum in 5 GHz \cite{LTEWiFisurvey}.

Previous researches have proposed some solutions to advance the spectrum sharing efficiency. Kota in \cite{kota} and Sodagari et al. in \cite{exmethod_6503914} proposed a joint design for multi-input multi-output communication systems and radar to minimize the co-channel interference. The multi-objective loss functions were defined for all channel users to find the jointly optimal waveform. However, these optimization methods only work for specific configurations, and need to start over for every condition change or compute all potential configurations in advance and memorize them. Furthermore, they need all information available, including number of spectrum users, user types, or bandwidth allocation, etc., which is impractical for real-world applications. In the last decade, reinforcement learning (RL) has been a popular solution to spectrum sharing problem. Q-learning was applied in \cite{Q_8645080} to intelligently manage transmitting power of radar and communication systems for a joint radar-communication receiver. \cite{Q_fairCoex,Q_sensors,8288850} discussed the coexistence between the LTE-U and Wi-Fi networks, adopting conventional Q-learning methods to learn the optimal active duration in duty-cycle for LTE-U users to maximize throughputs while keeping fair access rights. The author of \cite{RL_8378616} formulated the radar tracking problem as Markov decision process and utilized policy iteration to search the optimal linear frequency modulation for different target conditions and interference. Beside above methods, an analytical model was proposed in \cite{9093212} to evaluate the throughput of Wi-Fi and LTE-LAA networks and multi-armed bandit algorithm was utilized to tune the contention window for throughput maximization subject to fairness constraint. These works require a full picture about the whole spectrum and maintain predefined Q-table which includes all possible state-action combinations; the learned policies usually struggle in stochastic and non-stationary spectrum dynamics, and face troubles when applying to complicated problems. On the other hand, some other solutions are based on partially-observable environments. \cite{6482133,6417543,6415750,7249182} considered partially-observable Markov decision processes for wireless transceivers and sought for maximizing the transmitting efficiency in noisy spectrum. Gaussian process was adopted in \cite{9064881} for a time-series POMDP model to approximate each Q function in Q-table in consideration of correlation between channels. Author of \cite{8646702} proposed dynamic Q dictionary which allowed adding new state-action pair during training. Secondary users of cognitive radio networks in \cite{Q_7925694,Q_5986378,Q_dynchanselCR} utilized Q-learning to find the optimal policies for locating the clear bands in different spectrum configurations. Except these, deep learning is also grasping attentions. A model-free decentralized deep Q-learning method was combined with model-based Q-learning in \cite{9069218} to compensate the imperfectness of each other while accelerating the learning rate. Although deep Q-learning can handle partial observability, it still needs a pre-defined Q-table and approximates each Q function with a neural network (NN), which means a significant number of NNs and each NN requires a great bunch of data to train, let alone extra regularization in avoidance of overfitting.
\section{Bayesian Nonparametric Model}
\label{sec: BNP}
Bayesian statistics exploits the prior belief and provides an update of our belief about the unknown variables in a model using samples from the model. Denote $z$ as the desired variable in the model of interest, and $\mathcal{D}=(x_1,x_2,...,x_N)$ as the data set with each element drawn from the model, the Bayesian update for the belief about $z$ is represented through inverse probability rule
\begin{equation}
    p(z|\mathcal{D})=\frac{p(\mathcal{D}|z)p(z)}{p(\mathcal{D})}
\end{equation}
The main difference that distinguishes Bayesian inference from point estimation techniques like maximum likelihood is that Bayesian approaches regard unknown value $z$ as random variable. The prior $p(z)$ is the belief about how the value of $z$ will distribute before observing the first sample. The distribution $p(z)$ encodes all information known to us about the value of $z$ a priori. Since each $z$ value defines a unique estimate of the true model, the distribution over $z$ determines the inference over the true model. Once sample $(x_1,x_2,...)$ is collected from the true model, we can utilize it to update the belief to the posterior $p(z|\mathcal{D})$ by applying Bayes rule. $p(\mathcal{D}|z)$ is represented by the distribution family (usually a parametric model) over true model and is controlled by $z$, indicating the likelihood of observing $\mathcal{D}$ given some $z$ drawn from $p(z)$; the denominator $p(\mathcal{D})$ is the marginal likelihood over $\mathcal{D}$ and is obtained by marginalizing $z$ from the numerator. 

The prior model in Bayesian inferences can be categorized into two classes: parametric and nonparametric models. Parametric models have fixed structure and are simple to understand. These models are often utilized when the structure of the distribution family over the true model is well-defined. Nonparametric models, on the other hand, generalize the parametric models to infinite dimension to address a wider range of problems. The idea of utilizing nonparametric model is to reserve flexibility of adjusting model structures. In general, parametric priors can work well when the structure of true model is simple and some critical information can be known a priori. However, strong prior assumption is imposed on the structure of models, which is not the case in most real-world applications. For example, in an inference problem for Gaussian mixture model (GMM), if the number of Gaussian components is known, the prior model can be clearly defined and the inference task can be accomplished very efficiently and accurately. But if such information is unavailable, parametric methods may need to perform inference many times, each with different setting about the number of components, and incorporate extra algorithms like cross validation to select the optimal setting, which causes the solution inefficient and burdensome. Nonparametric models, on the other hand, can solve such inference problem with single algorithm. Nonparametric models treat the structure of the model as extra variable, loosening the limitation of parametric models. By incorporating infinite possibility over model structure, nonparametric models allow the learning machine to learn model parameter and structure together from data, thus can apply to wider range of models with less prior information required.

Dirichlet process (DP) is a commonly-used nonparametric model in discrete cases. It is a generalization of Dirichlet distribution and is first proposed by Ferguson in 1973 \cite{ferguson1973}. Denote $G_0$ as a distribution over probability space $\Theta$ and $\alpha$ as some positive real value, $\theta_1, \theta_2,...$ are drawn i.i.d from $\Theta$ given $G_0$ with corresponding probability $p_1,p_2,...$, a random measure $G$ is represented as discrete distribution with infinitely countable components,
\begin{equation*}
    G=\sum_{i=1}^{\infty}p_i\delta_{\theta_i}, \quad \sum_{i=1}^{\infty}p_i=1,
\end{equation*}
where $\delta$ means the Dirac delta function. From definition, $G$ is distributed according to $\text{DP}(\alpha, G_0)$ if for arbitrary finite measurable partition $(A_1,...,A_n)$ over $\Theta$, the vector of random measure $G(A_1),...,G(A_n)$ follows Dirichlet distribution,
\begin{equation*}
    (G(A_1),...,G(A_n))\sim \text{Diri}(\alpha G_0(A_1),...,\alpha G_0(A_n))
\end{equation*}
Chinese restaurant process (CRP) \cite{aldous1985CRP} and stick-breaking (SB) process \cite{sethuraman1994SB} are two different metaphors for realization of $\text{DP}(\alpha, G_0)$, both incorporate unbounded process of generating samples. I particularly focus on two methods of constructing Dirichlet processes. 

\subsection{Chinese Restaurant Process}
\label{sec: CRP}

Chinese restaurant process defines random distribution over partitions of samples. Given a finite set of samples $x_1,...,x_N$ and an infinite set of clusters, first sample is assigned to the first cluster with probability $1$, and the $n$th sample is assigned to the $k$th non-empty cluster with probability proportional to $n_k$, the number of samples already in the cluster in previous $n-1$ samples, and to a new cluster with proportional to $\alpha$. Denote $\theta_n$ as the cluster parameter for $x_n$, and $(\theta_1^*,...,\theta_K^*)$ represents the unique set of cluster parameters i.i.d drawn from $G_0$ for previous $n-1$ samples, the CRP can be expressed as
\begin{equation}
\begin{aligned}
    & p(\theta_1 = \theta_1^*)=1 \\
    & (\theta_n|\theta_1,...\theta_{n-1})=
    \begin{cases}
    \theta_k^* & \text{with probability }\frac{n_k}{n-1+\alpha} \\
    G_0 & \text{with probability }\frac{\alpha}{n-1+\alpha}
    \end{cases}
\end{aligned}
\label{equ: CRP}
\end{equation}
It can also be written in equation
\begin{equation*}
    p(\theta_n|\theta_1,...\theta_{n-1})=\sum_{k=1}^{K}\frac{n_k}{n-1+\alpha}\delta_{\theta_k^*}+\frac{\alpha}{n-1+\alpha}G_0
\end{equation*}
De Finetti's theorem implies that the order of both the clusters and the samples in each cluster is exchangeable because of conditional independence given $G$, thus each sample can be placed at the last position so that it is conditional on all others. It is obvious that the value of $\alpha$ determines the increase of clusters. \ref{equ: CRP} exhibits that probability of cluster assignment only depends on the cluster size, which means the larger the $n_k$ is, the higher the probability to cluster $k$ is. This rich-gets-richer phenomenon helps govern the growth of cluster number. 

\subsection{Stick-Breaking Process}
\label{sec: SBP}
\begin{figure}
    \centering
    \begin{subfigure}[b]{0.49\textwidth}
        \centering
        \includegraphics[width=\textwidth]{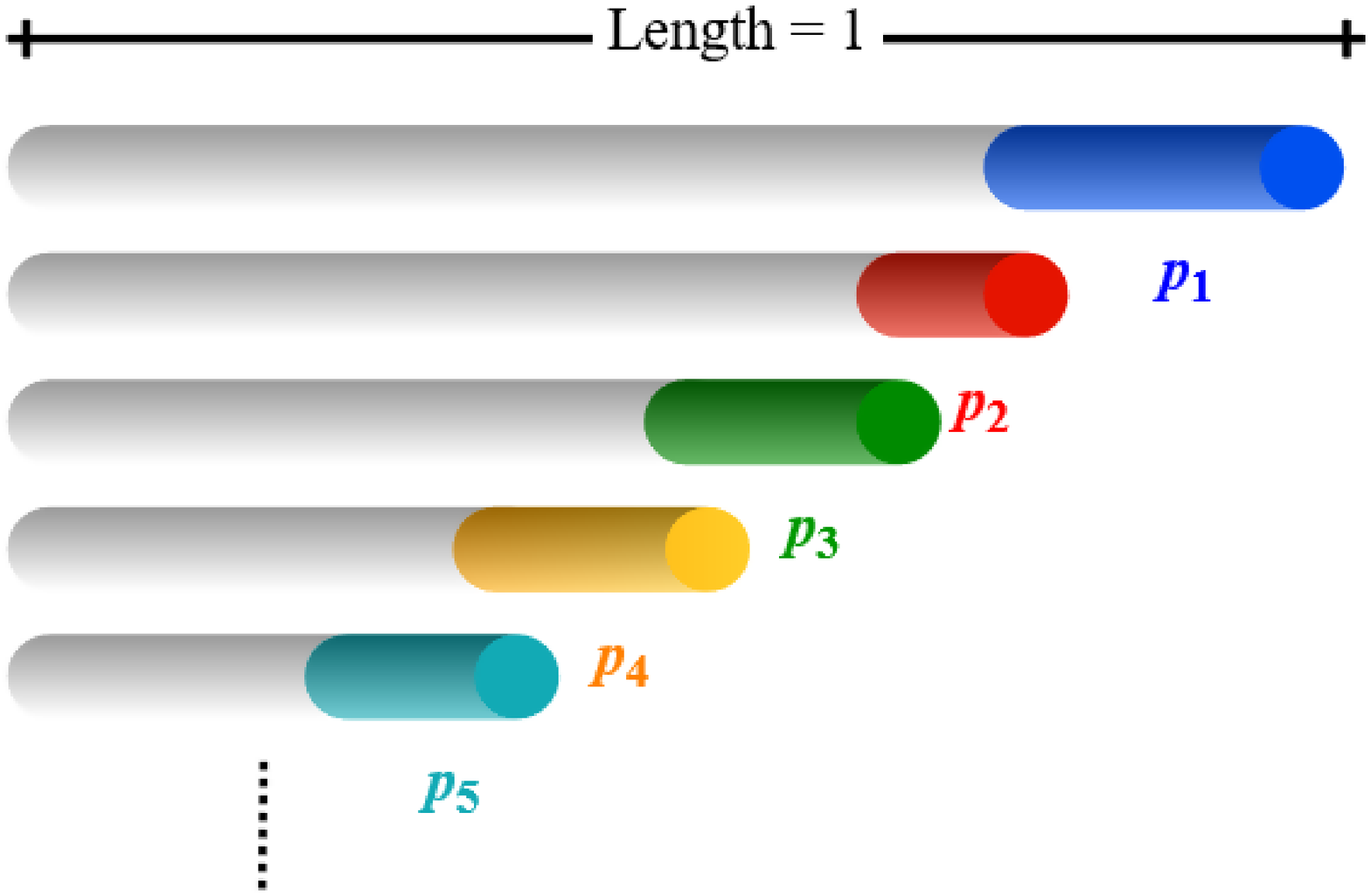}
        \caption[]{{\small Visualization}}
        \label{fig:SB_visual}
    \end{subfigure}
    \hfill
    \begin{subfigure}[b]{0.49\textwidth}  
        \centering 
        \includegraphics[width=\textwidth]{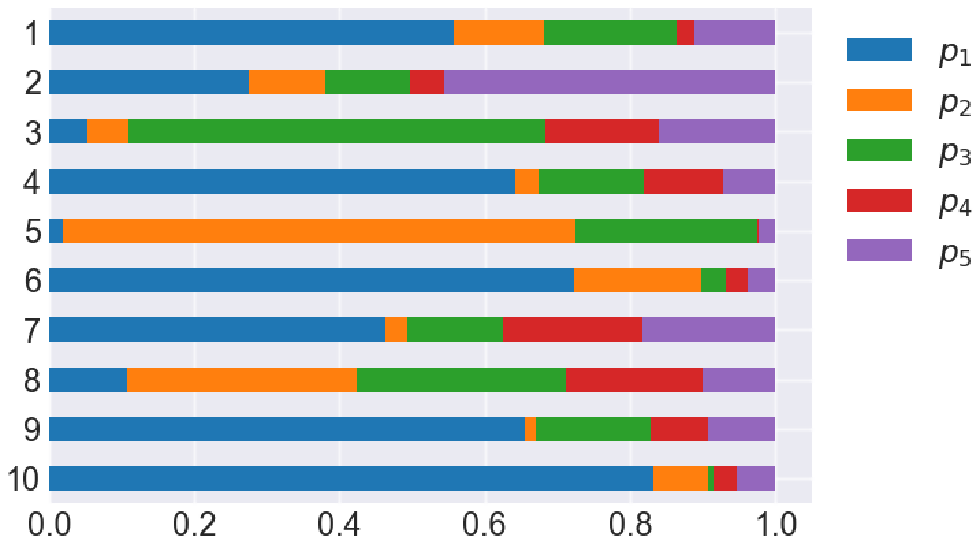}
        \caption[]{{\small Simulation with $\alpha=2$}}
        \label{fig:SB_simulate}
    \end{subfigure}
\caption[Illustration of The Stick-Breaking Process With Simulation]{Illustration of the stick-breaking process with simulation. (a) Visualization of the infinite process of breaking a unit-length stick into pieces, (b) Simulation for the stick-breaking process; the process is truncated at 5. All values of breaking portion are generated with $\alpha=2$.}
\label{fig:stick-breaking}
\end{figure}

\begin{figure}
\centerline{\includegraphics{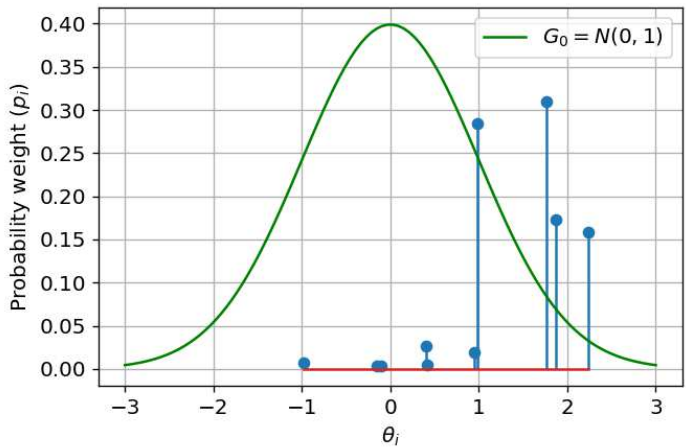}}
	\caption[SB Construction for DP]{A stick-breaking construction for $G$ with base distribution as standard normal distribution and $\alpha=2$.}
	\label{fig:SB}
\end{figure}

The stick-breaking process provides a straightforward approach to construct $G$. The approach to generate the probability weights of $G$ in stick-breaking process is analogous to breaking a unit-length stick into infinite number of pieces. Consider a stick with length 1 initially, we first break a portion $V_1$ off from the stick. As the process proceeds, at time $i$ a portion $V_i$ will be broken off from the remaining part of the stick. The values of the breaking portion are determined by Beta distribution. \ref{fig:SB_visual} visualizes this process with simulation results. Given a random variable $V$ with beta distribution $\text{Beta}(1,\alpha)$, and point mass $\theta_1,\theta_2,...$ drawn from $G_0$, the random probability weights $(p_1,p_2,...)$ in $G$ can be construct through an unbounded process:
\begin{equation}
    \begin{aligned}
    &\begin{cases}
    \theta_i|G_0\myeq G_0 \\
    V_i|\alpha \myeq \text{Beta}(1,\alpha)
    \end{cases}, i = 1,2,... \\
    & p_1 = V_1, \  p_i= V_i\prod_{j=1}^{i-1}(1-V_j) \text{ for }i>1 \\
    & G=\sum_{i=1}^{\infty}p_i\delta_{\theta_i}
    \end{aligned}
\label{equ: SB}
\end{equation}
\ref{fig:SB_simulate} demonstrates different simulation results for $\alpha=2$. \ref{fig:SB} illustrates an example of stick-breaking construction for $G$ with standard normal distribution as $G_0$ and $\text{Beta}(1,0.2)$ for generating the probability weights $p_i$ (in this example, the stick-breaking process is bounded for simplicity, but it can proceed to infinity). The stick-breaking process can construct the random measure $G$ fast and guarantee the probabilities $p_i$ sum to $1$. The distribution over $p_i$ is also known as $\text{GEM}(\alpha)$ distribution \cite{pitman2002GEM}.
\subsection{Application of Bayesian Nonparametric Model}

Bayesian nonparametric modeling has been adopted in signal processing, especially when the data pattern is uncertain a priori. In \cite{DBLPBahman, moraffah2019random, moraffah2018dependent, moraffah2019inference, moraffah2019nonparametric, moraffah2019tracking,moraffah2021clutter} the author proposed a dependent Dirichlet process to infer the unbounded number of objects and characteristics of each object together in a radar tracking scenario. Hierarchical Dirichlet process is also used in tracking multiple time-varying objects \cite{moraffah2019use, moraffah2019inference}.  Guo et al. in \cite{guo2018explaining} suggested that the Dirichlet process mixture model can be utilized to extract insights from deep neural networks, assisting understanding and interpretation of machine learning models, and demonstrating its ability to generalize to different machine learning frameworks. A variance Gamma process was proposed in \cite{pmlr-zhang18j} to encode probabilistic assumptions in the model prior, interpreting sparse and discrete data points in time-series data better than traditional machine learning algorithms, which usually yield smooth function. The author of \cite{DBLPPolatkan} adopted Beta-Bernoulli process over the model prior to learn an unbounded set of visual recurring patterns from data, and utilized this learned set to augment image resolution from low-resolution images.
\section{Sampling Algorithm}
\label{sec: sampling}
A central task in the application of probabilistic inference is the evaluation of the posterior distribution $p(z|\mathcal{D})$, and consequently computing the expectations with respect to the desired distribution. In Bayesian inference, the most straightforward way to obtain the information from the posterior is Monte Carlo method. Monte Carlo methods directly draw independent samples from the posterior distribution. However, drawing independent samples from the posterior distribution is not always possible. Markov chain Monte Carlo (MCMC) is a general framework for drawing dependent samples from various distributions. Monte Carlo methods, in general, estimate the statistical property of the desired distribution from random samples generated from it; Markov chain property means that samples are randomly generated by a sequence of process, where each samples is drawn from the conditional distribution given only its previous sample (Markov property), which means
\begin{equation*}
    p(x_n|x_{n-1},...,x_{1})=p(x_n|x_{n-1})
\end{equation*}
\subsection{Gibbs Sampling}
\label{sec: gibbs}
One the widely used and well-understood MCMC algorithms in practice is Gibbs sampling. The main idea behind Gibbs sampling is that it adopts iterative procedure to drawing samples from the target distribution. Consider a joint posterior distribution of variables $p(\mathbf{z}|\mathcal{D})=p(z_1,...,z_K|\mathcal{D})$, the objective is to drawing samples $(\mathbf{z}^1,\mathbf{z}^2,...,\mathbf{z}^t,...)$, where each $\mathbf{z}^t$ is a vector of samples $(z_1^t,...,z_K^t)$. At $t$th Markov state, Gibbs sampling sequentially draws sample of each $z_i^t$ conditioned on the latest sampled values of all other variables, that is, sample of $z_i^t$ is generated from the conditional distribution $p(z_i|z_1^t,...,z_{i-1}^t,z_{i+1}^{t-1},...,z_{K}^{t-1},\mathcal{D})$. The sample of next variable $z_{i+1}^t$ is then drawn from the distribution given the sampled value of $z_{i}^t$. Once all variables are sampled at $t$th state, the sample vector $\mathbf{z}^t$ is complete and iteration proceeds to next state. The process is shown in the following

\begin{algorithm}[H]
\SetKwInOut{Input}{Input}\SetKwInOut{Output}{Output}\SetKwInOut{Initialize}{Initialize}\SetKw{Return}{Return}
\SetAlgoLined
\Initialize{values of $z_i^0$, $i=1,...,K$}
 \For{$t=1,...,T$}{
   sample $z_{1}^t\sim p(z_1|z_{2}^{t-1},...,z_{K}^{t-1},\mathcal{D})$\\
   sample $z_{2}^t\sim p(z_2|z_{1}^t,z_{3}^{t-1},...,z_{K}^{t-1},\mathcal{D})$\\
   $\vdots$\\
   sample $z_{i}^t\sim p(z_i|z_{1}^t,...,z_{i-1}^t,z_{i+1}^{t-1},...,z_{K}^{t-1},\mathcal{D})$\\
   $\vdots$\\
   sample $z_{K}^t\sim p(z_K|z_{1}^t,...,z_{K-1}^t,\mathcal{D})$\\
   construct $\mathbf{z}^t=(z_{1}^t,...,z_{K}^t)$
   }
\Output{sample sequence $(\mathbf{z}^1,...,\mathbf{z}^T)$}
 \caption{Gibbs sampling for $p(\mathbf{z}|\mathcal{D})$}
\label{alg: gibbs}
\end{algorithm}

\section{Variational Inference}
\label{sec: VI}
MCMC methods provide straightforward and feasible solution to approximate the exact models. However, there are some drawbacks in practical applications when the model complexity is increasing. First, it can be difficult to predict when the stochastic process converges. How converged results deviates from the true models is also difficult to quantify due to stochastic nature. Stochastic sampling means the statistical properties extracted from those samples usually require a lot of time and samples to reach stationary state, which means large storage space is necessary. Second, the amount of computation can increase exponentially when new variables are introduced to model. Such computationally demanding nature often refrains them from scaling to problems with high dimension. Finally, Gibbs sampling requires the conditional distribution to be analytical or the sampling from it will be complicated. Although stochastic approach can yield theoretically most accurate results given infinite computational resource, in practice only approximate estimates can be obtained due to finite amount of time and samples, which means the accuracy is determined by the limit of computational resource, and may never reach its theoretically optimum. 

In contrast, variational inference (VI) provides a useful alternative to compensate the drawbacks of sampling method. First, unlike directly sampling from the exact models, VI are approximate models, which serves as surrogate to estimate the statistical properties of true model. Second, its computation requires less computational resource than sampling method so is easier to apply to large problems. The derivation is deterministic and easy to measure how approximate the surrogate is to the true model. The result is guaranteed to be the optimal possible approximate to the objective models in its distribution family. Finally, it formulates the derivation of unknown distribution into an optimization problem so that it is convenient to apply many optimization techniques to improve the computation.

The core concept of VI is to utilize variational distribution $q(\mathbf{z})$ as surrogate to approximate the true posterior distribution $p(\mathbf{z}|\mathcal{D})$. The VI utilizes Kullback–Leibler divergence (KL-divergence) to quantify the deviation of $q(\mathbf{z})$ from $p(\mathbf{z}|\mathcal{D})$, which is defined as \cite{VI_review}:
\begin{equation}
    \text{KL}(q\|p)\coloneqq\int_\mathbf{Z} q(\mathbf{z})\ln\frac{q(\mathbf{z})}{p(\mathbf{z}|\mathcal{D})}d\mathbf{z}=\text{E}_q\left[\ln\frac{q(\mathbf{z})}{p(\mathbf{z}|\mathcal{D})}\right]
\label{equ: KL}
\end{equation}
If $q(\mathbf{z})$ is defined discrete distribution, the integration is then replaced by summation. The problem of finding a distribution is thus reformulated as an optimization problem which seeks an optimal $q^*(\mathbf{z})$ such that 
\begin{equation*}
    q^*(\mathbf{z})=\argmin_{q(\mathbf{z})}\text{KL}(q\|p)
\end{equation*}
Decompose \ref{equ: KL}, we get
\begin{equation}
\begin{aligned}
    \text{KL}(q\|p)&=\text{E}_q\left[\ln q(\mathbf{z})\right]-\text{E}_q\left[\ln p(\mathbf{z}|\mathcal{D})\right] \\
    &=\text{E}_q\left[\ln q(\mathbf{z})\right]-\text{E}_q\left[\ln \frac{p(\mathbf{z},\mathcal{D})}{p(\mathcal{D})}\right] \\
    &=\text{E}_q\left[\ln q(\mathbf{z})\right]-\text{E}_q\left[\ln p(\mathbf{z},\mathcal{D})\right]+\text{E}_q\left[\ln p(\mathcal{D})\right] \\
    &=\text{E}_q\left[\ln q(\mathbf{z})\right]-\text{E}_q\left[\ln p(\mathbf{z},\mathcal{D})\right]+\ln p(\mathcal{D})
\end{aligned}
\label{equ: KL2}
\end{equation}
$\ln p(\mathcal{D})$ is the log marginal likelihood of data set $\mathcal{D}$ and has nothing to do with variable set $\mathbf{z}$, so the expectation over it is not functioning. Rearrange \ref{equ: KL2} we obtain
\begin{equation}
\begin{aligned}
    \ln p(\mathcal{D})&=\text{KL}(q\|p)+\text{E}_q\left[\ln p(\mathbf{z},\mathcal{D})\right]-\text{E}_q\left[\ln q(\mathbf{z})\right] \\
    &=\text{KL}(q\|p)+\text{ELBO}(q)
\end{aligned}
\label{equ: ELBO}
\end{equation}
The term $\text{ELBO}(q)$ is the evidence lower bound of $q$ distribution. Since the left-hand side of \ref{equ: ELBO} is constant, the $q(\mathbf{z})$ that minimizes $\text{KL}(q\|p)$ is just the one that maximizes $\text{ELBO}(q)$. If we maximize $\text{ELBO}(q)$ by optimization with unrestricted choices of $q(\mathbf{z})$, the maximum value of $\text{ELBO}(q)$ happens when $\text{KL}(q\|p)$ is equal to $0$, which means the resulting $q^*(\mathbf{z})$ is just the objective distribution $p(\mathbf{z}|\mathcal{D})$. However, such setting will cause the problem intractable. To guarantee the optimization to converge, it is necessary to place some assumption on the form of $q(\mathbf{z})$. A commonly applied assumption is the mean-field approximation. Suppose we partition the variable set $\mathbf{z}$ into $K$ disjoint groups, and each group contains at least $1$ variable. Denote each group as $z_i$ where $i = 1, 2, ..., K$, we then assume the joint distribution $q(\mathbf{z})$ can factorize into the product of all $q(z_i)$s,
\begin{equation}
    q(\mathbf{z})=\prod_{i=1}^{K}q(z_i).
\label{equ: meanfield}
\end{equation}

By decomposing $q(\mathbf{z})$ into the product of independent marginal $q(z_i)$s, we then can maximize $\text{ELBO}(q)$ with respect to each $q(z_i)$ individually and multiply them to obtain the joint $q(\mathbf{z})$. It is necessary to emphasize that the mean-filed approximation is the only assumption we place on $q$ distribution. The process of optimization turns to seek the $q^*(\mathbf{z})$ which satisfies \ref{equ: meanfield} with maximum $\text{ELBO}(q)$ value. Substitute \ref{equ: meanfield} into the form of $\text{ELBO}(q)$, we obtain
\begin{equation}
\medmuskip=1mu
\thinmuskip=1mu
\thickmuskip=1mu
\begin{aligned}
    &\text{ELBO}(q) \\
    &=\text{E}_q\left[\ln p(\mathbf{z},\mathcal{D})\right]-\text{E}_q\left[\ln \prod_{i=1}^{K}q(z_i)\right] \\
    &=\text{E}_q\left[\ln p(\mathbf{z},\mathcal{D})\right]-\sum_{i=1}^{K}\text{E}_q\left[\ln q(z_i)\right] \\
    &=\int q(\mathbf{z})\ln p(\mathbf{z},\mathcal{D})d\mathbf{z}-\sum_{i=1}^{K}\int q(\mathbf{z})\ln q(z_i)d\mathbf{z} \\
    &=\iint q(z_i)q(\mathbf{z}_{-i})\ln p(\mathbf{z},\mathcal{D})d\mathbf{z}_{-i}d z_i-\sum_{i=1}^{K}\iint q(z_i)q(\mathbf{z}_{-i})\ln q(z_i)d\mathbf{z}_{-i}d z_i \\
    &=\int q(z_i)\left[\int q(\mathbf{z}_{-i})\ln p(\mathbf{z},\mathcal{D})d\mathbf{z}_{-i}\right]d z_i -\sum_{i=1}^{K}\int q(z_i)\ln q(z_i)\left[\int q(\mathbf{z}_{-i})d\mathbf{z}_{-i}\right]d z_i \\
    &=\int q(z_i)\text{E}_{q_{-i}}\left[\ln p(\mathbf{z},\mathcal{D})\right] d z_i-\sum_{i=1}^{K}\int q(z_i)\ln q(z_i)d z_i \\
    &=\int q(z_i)\text{E}_{q_{-i}}\left[\ln p(\mathbf{z},\mathcal{D})\right] d z_i-\int q(z_i)\ln q(z_i)d z_i-\sum_{j\neq i}\int q(z_j)\ln q(z_j)d z_j
\end{aligned}
\label{equ: opt_elbo}
\end{equation}
$q(\mathbf{z}_{-i})=\prod_{j\neq i,\, j=1}^{K}q(z_j)$ is the joint distribution of all variable groups other than $z_i$. To maximize $\text{ELBO}(q)$ with respect to some $q(z_i)$, we take partial derivative on \ref{equ: opt_elbo} with respect to $q(z_i)$ and set it equal to 0, subject to the condition that $q(z_i)$ must integrate to 1,
\begin{equation}
\begin{aligned}
    &\frac{\partial\text{ELBO}(q)}{\partial q(z_i)}=\text{E}_{q_{-i}}\left[\ln p(\mathbf{z},\mathcal{D})\right]-\ln q(z_i)-1=0 \\
    &\rightarrow \ln q(z_i)=\text{E}_{q_{-i}}\left[\ln p(\mathbf{z},\mathcal{D})\right]-1 \\
    &\rightarrow q^*(z_i)\propto \exp\left\{\text{E}_{q_{-i}}\left[\ln p(\mathbf{z},\mathcal{D})\right]\right\}
\end{aligned}
\label{equ: VIoptformula}
\end{equation}

This formula exhibits what optimal $q^*(z_i)$ would look like. In general, we do not specify the form of $q(z_i)$ a priori, however, if prior distribution $p(z_i)$ and likelihood $p(\mathcal{D}|z_i)$ have conjugacy, the approximation $q^*(z_i)$ to the posterior $p(z_i|\mathcal{D})$ inherently shares the same form of $p(z_i)$. This is useful in computing each $q(z_i)$ analytically when $p(\mathbf{z},\mathcal{D})$ is well-defined. and A commonly utilized algorithm to optimize each $q(z_i)$ is the coordinate ascent variational inference (CAVI). CAVI iteratively optimizes each $q(z_i)$, while keeping all others fixed. It guarantees the $\text{ELBO}(q)$ to converge to local maximum. This algorithm is presented as follows:

\begin{algorithm}[H]
\SetKwInOut{Input}{Input}\SetKwInOut{Output}{Output}\SetKwInOut{Initialize}{Initialize}\SetKw{Return}{Return}
\SetAlgoLined
\Input{data set $\mathcal{D}$ and $p(\mathbf{z},\mathcal{D})$}
\Initialize{each $q(z_i)$ with respective initial parameters}
 \While{$\textsc{ELBO}(q)$ not converged \textsc{OR} iteration $<$ max}{
  \For{$i=1,...,K$}{
   determine $q^*(z_i)\propto \exp\left\{\text{E}_{q_{-i}}\left[\ln p(\mathbf{z},\mathcal{D})\right]\right\}$
   }
   compute $\text{ELBO}(q)$
 }
\Output{variational distribution $q(\mathbf{z})=\prod_{i=1}^{K}q(z_i)$}
 \caption{Coordinate Ascent Variational Inference}
\label{alg: cavi}
\end{algorithm}

On the other hand, the proxy $q(\mathbf{z})$ can estimate the mean of the true posterior $p(\mathbf{z}|\mathcal{D})$ accurately, but tend to underestimate the variance. Mean-field assumption simplifies the computation by dismissing potential dependency between variables, so the joint $q(\mathbf{z})$ can perform well but the marginal $q(z_i)$ may not. Conjugacy is not required in VI, but extra variables may be needed to govern $q(\mathbf{z})$ for non-conjugate cases, which could cause the problem intractable.

\section{(Partially-Observable) Markov Decision Process}

Markov decision process describes the interaction between agent and a (stochastic) environment. A typical Markov decision process comprises a tuple $\langle n, \mathcal{A}, \mathcal{S}, T, R, \gamma\rangle$, where each element represents one component of Markov decision process. $n$ is the agent of Markov decision process. An agent is a decision maker that can determine what action to perform based on its current state from the environment and receive feedback from the environment. In control system, it can be regarded as input generator which provides input to a system given the readings of the system. $\mathcal{A}$ represents the action set; the agent determines an action $a\in\mathcal{A}$ to perform. By performing action, the agent receives feedback from the environment and observes a new state. The state set $\mathcal{S}$ is utilized to describe the dynamic of the environment. At each time moment, a state $s\in\mathcal{S}$ is a variable to demonstrate the current configuration of the the environment. At each time an action is performed, the state will transit to another one with some probability. $T(s',a,s)=\Pr(s'|a,s)$ $\forall s,s'\in\mathcal{S}$, $a\in\mathcal{A}$ denotes the state transition probability given current state and action. Markov decision process assumes Markov property for the state transition, which means the distribution over state at time $t$ only depends on the state at $t-1$, that is, previous one state encompasses all information of the past state transition history.
\begin{equation*}
    \Pr(s_{t}|s_{t-1},s_{t-2},...)=\Pr(s_{t}|s_{t-1})
\end{equation*}
$R:\mathcal{S}\times\mathcal{A}\rightarrow\mathbb{Z}$ is the immediate reward function which feeds back a real value $r$ to the agent for every $(s, a)$ pair. It is worth to note that the design of reward function encodes the core objective in the learning process, that is, what is the prior concern for the agent; different reward functions will guide the learning process to different results. To avoid divergence of the learning process, discount factor $\gamma$ is introduced, which is a predefined positive real constant between $[0,1)$ utilized to reflect the importance of future rewards in contrast to the current one.

When the states of the environment are not fully-observable to the agent, the Markov decision process transforms to partially-observable. In such case, a partially-observable Markov decision process can be described by the tuple $\langle n, \mathcal{A}, \mathcal{S}, \mathcal{O}, T, \Omega, R, \gamma\rangle$. $n$, $\mathcal{A}$, $\mathcal{S}$, $T$, $R$, and $\gamma$ are the same as MDP model. But for POMDP model, an observation $o\in\mathcal{O}$ will be observed by the agent instead of $s$ after each action is performed. Each $o$ carries partial information about the true global state, and the observation function $\Omega(o)=\Pr(o|s,a)$ describes the probability distribution over observations when performing action $a$ and arriving at state $s$ at each time.

\subsection{Decentralized Partially-Observable Markov Decision Process}
\label{sec: decPOMDP}

When POMDP model generalizes to multi-agent scenario with each agent executes its own reinforcement learning without cooperation or information exchange, it becomes a decentralized POMDP model. A Dec-POMDP model can be represented by $\langle\mathcal{N}, \mathcal{A}, \mathcal{S}, \mathcal{O}, T, \Omega, R, \gamma\rangle$ \cite{POMDPbook,RLbook}. $\mathcal{S}$ and $\gamma$ are identical to POMDP model, and $\mathcal{N}$, $\mathcal{A}$, and $\mathcal{O}$ generalize to multi-agent case. $\mathcal{N}={1,...,N}$ is the finite set of agents. $\mathcal{A}=\bigotimes_n\mathcal{A}_n$ and $\mathcal{O}=\bigotimes_n\mathcal{O}_n$ correspond to the sets of joint actions and observations, where $\mathcal{A}_n$ and $\mathcal{O}_n$ are local action and observation sets for agent $n$. At each state, a joint action $\vec{a}=\left\{a_n\right\}_{n=1}^N \in\mathcal{A}$ is formed by the local actions $a_n\in\mathcal{A}_n$, and Joint observation $\vec{o}=\left\{o_n\right\}_{n=1}^N \in\mathcal{O}$, where $o_n$ is only accessible to agent $n$. $T$, $R$, and $\Omega$ are now functions of the joint action and observation. $T(s',\vec{a},s)=\Pr(s'|\vec{a},s)$ $\forall s,s'\in\mathcal{S}$, $\vec{a}\in\mathcal{A}$, $\Omega(\vec{o})=\Pr(\vec{o}|s,\vec{a})$, and $R$ is the global immediate reward function which yields rewards $r=R(s,\vec{a})$ for all agents.

\section{Reinforcement Learning}
\begin{figure}
\centerline{\includegraphics[width=0.8\linewidth]{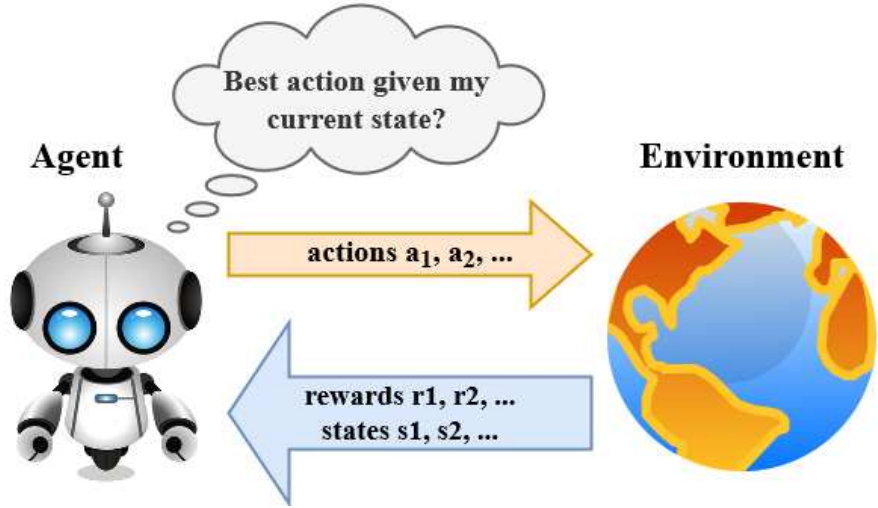}}
	\caption[Reinforcement Learning in MDP Environment]{The process of reinforcement learning is to find an action decision strategy given states which rewards the agent the most in the long run.}
	\label{fig:RL_basic}
\end{figure}
In a (PO)MDP environment, a policy $\pi$ is a function mapping current state/observation to a probability distribution over actions. 
\begin{equation*}
    \pi(s_{t})=\Pr(a_{t}|s_{t})
\end{equation*}
As we mentioned above, the (PO)MDP feeds back an immediate reward for every action performed, as time proceeds, the agent collects all received rewards. The value of a policy $\pi$ at each state is evaluated by the Bellman equation $V_{\pi}$, which is the expected sum of discounted future rewards for an amount of time with respect to the policy. 
\begin{equation*}
     V_{\pi}(s_t)=\text{E}_{\pi}\left[\sum_{t}\gamma^t R(s_t,a_t)\right]
\end{equation*}
The optimal policy $\pi^*$ is defined as the one which yields the maximal value at every state.
\begin{equation*}
     V_{\pi^*}(s_t)=\max_{\pi}\text{E}_{\pi}\left[\sum_{t}\gamma^t R(s_t,a_t)\right], \, \forall s_t\in\mathcal{S}
\end{equation*}
\ref{fig:RL_basic} illustrates the fundamental about how reinforcement learning works. At each time moment, the agent will select an action to perform to the environment given the current state the agent has observed, a real-valued reward will be received from the environment as feedback for the action; then the time index proceeds to next one. The state of the environment at next moment may change due to the action and will be observed by the agent. The above procedure is termed as an interaction. By interacting with the (PO)MDP environment many times, the agent gathers rewards and gradually learns a decision-making strategy, which suggests the agent how much the environment will reward the agent in the long run for the action it selects given current state. Reinforcement learning is thus the process of learning the optimal decision-making strategy, i.e., the policy, that will yield the most reward for underlying (PO)MDP model. Reinforcement learning is categorized as unsupervised learning, which has no correct answer to compare with during learning. Unlike other types of machine learning, the data for learning is not provided a priori but collected by interaction between agent and environment in each learning iteration. In Dec-POMDP model, due to lack of cooperation, each agent maintains its local policy $\pi_n$ which maps local observation history to local actions. All local policies constitute the joint policy. For all agent, the objective is to figure out a joint policy $\Pi=\bigotimes_n \pi_n$ that maximizes the long-term value function.
\subsection{Bayesian Reinforcement Learning}
\label{sec: BRL}
Bayesian reinforcement learning applies Bayesian inference to the values to be estimated in reinforcement learning, placing prior model over the desired values and infer the posterior model. In this work we adopt policy-based learning, which learns policy directly without knowing the underlying environment. Thus the desired values are policy parameters. Denote $\Theta$ the parameters of policy and $\mathcal{D}$ the data collected from interaction with the environment, the Bayesian policy learning infers the posterior model from the prior and data:
\begin{equation*}
    p(\Theta|\mathcal{D})=\frac{p(\mathcal{D}|\Theta)p(\Theta)}{p(\mathcal{D})}\propto p(\mathcal{D}|\Theta)p(\Theta).
\label{equ: bayes}
\end{equation*}
$p(\Theta)$ is the prior model indicating belief about $\Theta$ before observing the first datum. By applying distribution over $\Theta$, it is easy to encode auxiliary constraints to avoid undesired results, and quantify our confidence about the value of $\Theta$. Through interaction with the environment, data is collected and utilized to derive likelihood $p(\mathcal{D}|\Theta)$ to infer the posterior $p(\Theta|\mathcal{D})$. In iterative reinforcement learning, the posterior distribution obtained at current iteration can serve as the prior distribution at next iteration. By performing the iteration many times, the convergence of $p(\Theta)$ can be guaranteed. Bayesian learning provides a faster and simpler comparison to deep learning since the presence of prior model provides a bias to the learned model to avoid overfitting in nature so that extra regularization is not needed; prior knowledge encoded in prior models also mitigates the desire for data to obtain the matching performance as deep learning.
\chapter{BAYESIAN REINFORCEMENT LEARNING IN COEXISTENCE}
\label{chap:BRL_SS}
\acresetall
In this chapter the LTE and Wi-Fi coexistence mechanism in working IEEE specification is presented, then a Dec-POMDP model based on it is formulated, involving a cumulative reward function to reflect the continuous channel dynamics. Then the nonparametric models placed for prior distribution is proposed. We utilize variational inference to approximate the posterior distribution, the analytical approximation result will also be demonstrated.
\section{Problem Setup}
\begin{figure}
\centerline{\includegraphics[width=1\linewidth]{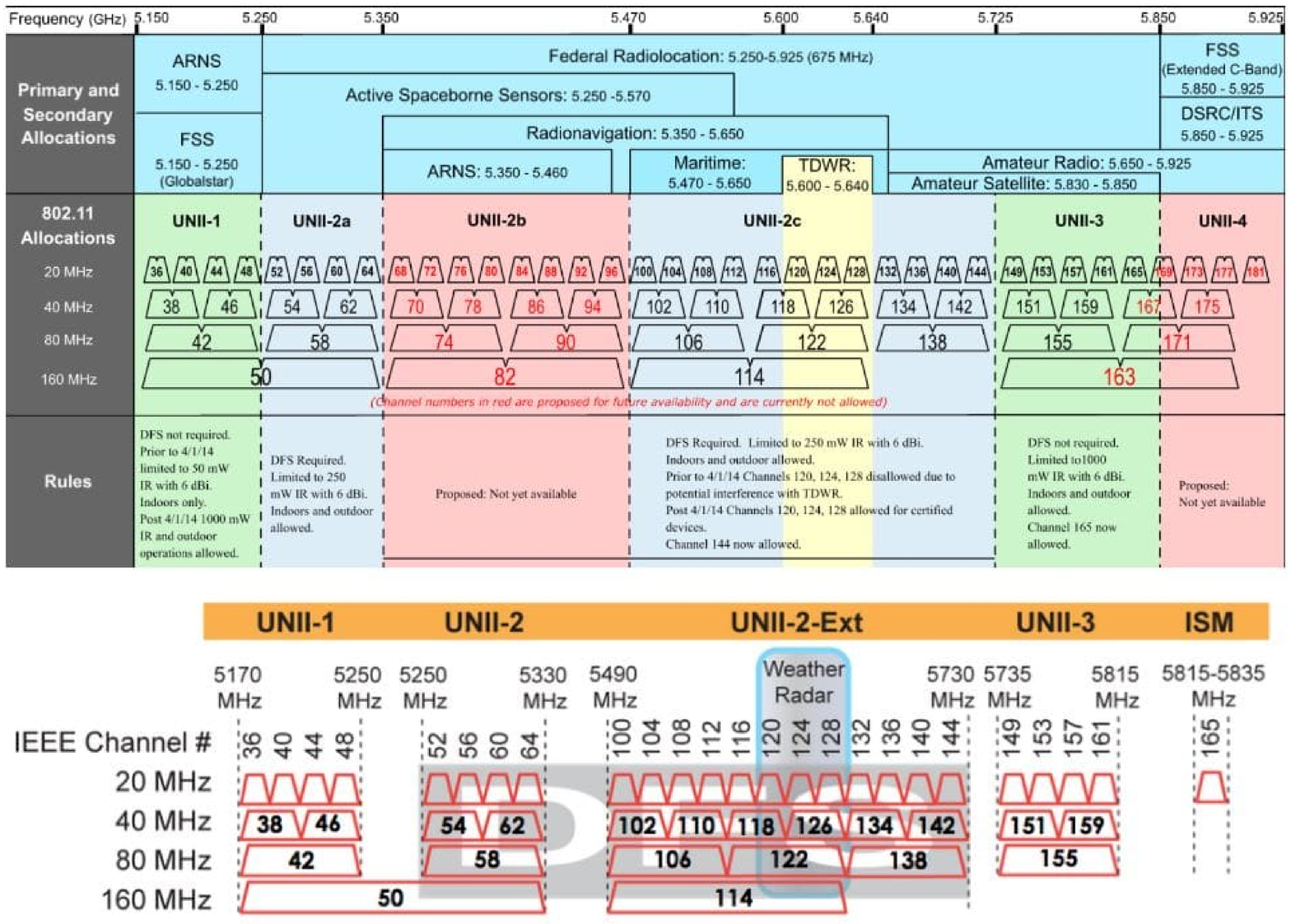}}
	\caption[5G Spectrum Usage]{Illustration for current frequency allocation in 5 GHz spectrum\footnotemark.}
	\label{fig:5GHz}
\end{figure}
%
The 5 GHz unlicensed band has approximately 600 MHz and is divided into non-overlapping channels. \ref{fig:5GHz} illustrates the current frequency allocation scenario in 5 GHz unlicensed spectrum. The minimum unit for channel allocation has bandwidth of 20 MHz. If the spectrum is not crowded, the wireless node is allowed to utilize channel with wider bandwidth (40, 80, or 160 MHz), which is formed by combining multiple consecutive 20-MHz channels. In unlicensed spectrum, a wireless node can be either LTE evolved node B (eNB) or Wi-Fi access point (AP). Due to the property of unrestricted access, it is infeasible to have a control center to manage all heterogeneous nodes that attempt to access the spectrum; information exchange between heterogeneous networks also suffers trouble because of their divergent protocols and the consideration of extra costs. Thus a practical spectrum sharing scheme should be decentralized, which means cooperation between nodes is minimized and each node learns its own spectrum accessing policy independently. Additionally, nodes like user equipment only possess limited spectrum sensing capability and obtain partial information about the spectrum. Without observing the global configuration, nodes determine what action to take based on the sufficient statistics of past observations and actions, which is termed as belief or decision state in some documents. Given aforementioned conditions, the spectrum dynamic can be described as a Dec-POMDP model.

\footnotetext{From \url{https://www.wlanpros.com/5ghz-frequency-allocations-2/}.}

It is worth to note that the unbounded possibility of policy should be considered. The license-free property allows every node to enter and leave the spectrum unrestrictedly, albeit each time there is only a finite set of nodes has the opportunity to occupy the spectrum. Hence we should not expect the number of potential nodes is bounded and known a priori. The policy learning for each node must considers interactions with uncertain number of coexisting nodes, thus nonparametric models which can accommodate infinite policy representations are more appropriate than parametric models. On the other hand, fair spectrum sharing is another factor which is crucial to the coexisting networks and worth more attention. The LTE data frame is constituted by sub-frames, where each sub-frame lasts 1ms. The number of sub-frames conveyed in one transmission is determined by the access priority of the node \cite{3gpp.36.213}. The Wi-Fi data frame, on the other hand, is packet-based. Each Wi-Fi transmission contains only one packet. The frame aggregation in IEEE 802.11n/ac, which enhances airtime efficiency by combining multiple packets in single transmission \cite{802.11n_cisco}, is not the case in our problem. The different composition of data frame makes LTE transmission a lasting channel occupation while Wi-Fi a short burst, which causes LTE nodes more easily dominates the time allocation and thus winner keeps winning, expelling Wi-Fi nodes from the spectrum. If only the global performance of the spectrum is considered, sometimes the learning process will tend to sacrifice vulnerable nodes to benefit powerful ones, which is what we want to avoid. In our algorithm, we incorporate the most commonly-utilized Jain's fairness indicator \cite{jain1984fairness} as a measure in the reward function to resolve the potential unfairness. The Jain's fairness indicator was initially proposed to evaluate the network performance thus it is a favorable choice for our model. By introducing the fairness factor to weigh the reward from each node, the usage balance between nodes can be secure.

\subsection{Signal Model}
\label{sec: sigmodel}

\begin{figure}
\centerline{\includegraphics{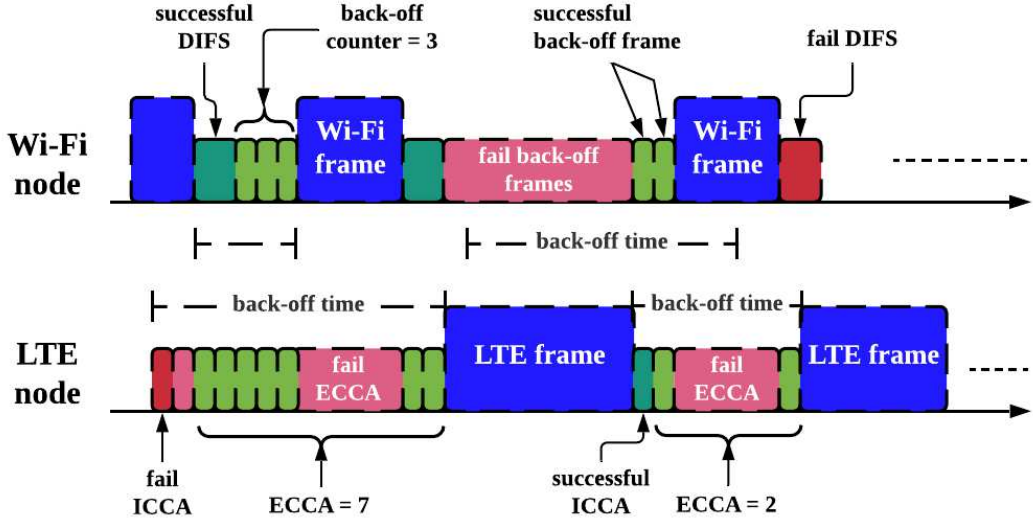}}
	\caption[LTE-LAA Spectrum Sharing]{Illustration of the spectrum sharing mechanism between LTE-LAA and Wi-Fi nodes.}
	\label{fig:lte-wifi}
\end{figure}

As we mentioned in \ref{sec: SS_litreview}, the Wi-Fi standard has been utilizing CSMA/CA for spectrum sharing among access points in the unlicensed spectrum. The CSMA/CA adopts sensing before transmission to avoid channel overload at a time. Before transmission starts through a channel, Wi-Fi nodes must perform an initial channel sensing for a Distributed Inter-Frame Spacing (DIFS) duration to evaluate channel status, access is suppressed if the channel is judged to be busy. If channel is sensed idle, Wi-Fi nodes then performs an additional back-off sensing to further inspect the channel status. During back-off sensing phase, a positive integer is generated randomly from a predefined range $[0, CW]$ as a down counter, where $CW$ means contention window. The counter counts down by $1$ for each time the channel is sensed idle in a fixed-length time slot. The countdown will freeze for any non-idle result and resume when the sensing result is idle again. The node has access to the channel once the counter reaches $0$. Stochastic back-off counters generated by different Wi-Fi nodes avoids collisions by staggering their access timing. Similar to W-Fi, The LTE-LAA standard enables LTE nodes to coexist with other nodes in unlicensed spectrum by implementing LBT mechanism. The main difference lies on the length of time slots in initial and back-off channel sensing phases. According to \cite{3gpp.36.213}, the $CW$ set and maximum allowed channel occupation time a LTE node can select depends on the channel access priorities. With larger $CW$ value, the LTE nodes are able to occupy the channel for a longer duration, so there is a trade-off between sensing duration and channel occupation time. The initial and back-off channel sensing mechanisms in LTE standard is termed as Initial Clear Carrier Assessment (ICCA) and Extended Clear Carrier Assessment (ECCA). It is important to note that in our model, the access priorities are equal for both Wi-Fi and LTE nodes, and the back-off sensing shall be performed by any means after the channel is judged as idle in the initial sensing phase. A simple example of our spectrum sensing scheme is illustrated in \ref{fig:lte-wifi}.

\subsection{Model Formulation}

Here we define each component of the Dec-POMDP model for our spectrum sharing scenario to apply reinforcement learning algorithm.
\begin{itemize}[leftmargin=*]
\item {\em Agents:} each agent in our framework is the network manager, which can be either a LTE-LAA eNB or Wi-Fi AP. There are $L$ number of LTE-LAA agents and $W$ number of Wi-Fi agents attempting to access the spectrum. Due to limited resource, only a subset of $N=L+W$ agents are able to access the spectrum at a time. We utilize notation $n$ for agent index.
\item {\em Actions:} for our wireless agents, each element $a_i$ in action set $\{a_1,a_2,...\}$ is a number representing the value of the contention window $CW$. Once an agent has selected an action $a_i$ from the action set, it will then sample an integer randomly from the region $\left[0, a_i\right]$ as its back-off counter. LTE and Wi-Fi agents share the same set of contention windows while the channel occupation time for LTE agents depends on the selected $CW$ value. All agents operate on the same channel, that is, frequency domain multiplexing is not our concern.
\item {\em States:} each global state $s_k$ corresponds to one spectrum configuration. A configuration is an integer which indicates the number of agents currently occupying the spectrum. There are total $(N+1)$ number of unique states.
\item {\em Observations:} in LTE-LAA and Wi-Fi standards demand agents to inspect the occupational status of the channel for additional time slots before transmission; the observation received by agent after each action is performed is defined as the duration between initial sensing starts to the end of back-off sensing, which is the time an agent actually spends in waiting for the channel resource, reflecting the occupation of the channel.
\item {\em Reward Function:} we want the reward function to reflect the influence from past history of actions and observations, thus the local reward is a cumulative function dependent of current and accumulation of past rewards. For wireless agents, it is desirable to exploit the channel resource as more efficient as possible. For each agent, the local reward is a function of the effective throughput $Th_n^t$ for agent $n$ at time $t$ reweighted by the Jain's fairness indicator. Denote $PL_n^t$ as the effective transmitted payload without colliding with any other transmission, and $D_n^t$ as the duration from initial channel sensing starts to transmission ends, the global reward received by nodes which complete their actions at time $t$ is defined in \ref{equ: rwd}.
\end{itemize}

\begin{equation}
\begin{aligned}
    & \text{Global reward }R_t = \sum_{n=1}^{N}r_t^n \\
    & \text{Local cumulative reward }r_t^n = r_{t-1}^n + R_n(t) \\
    & R_n(t) = \ln\left\{\left|\widetilde{Th}_{n}^t\right|+1\right\} \\
    & \widetilde{Th}_n^t = J_n^t Th_n^t, \ \ Th_n^t=\frac{PL_n^t}{D_n^t} \\
\end{aligned}
\label{equ: rwd}
\end{equation}
where $J_n^t$ is the Jain's fairness indicator \cite{jain1984fairness} and is computed by
\begin{equation}
J_n^t = \frac{\left(\sum_{\forall i\neq n}x_i^{t-1} + x_n^t\right)^2}{N\left(\sum_{\forall i\neq n}{x_i^{t-1}}^2 + {x_n^t}^2\right)}, \ \ x_i^t = \frac{Th_i^t}{O_i}, \ \forall i\in [1,N]
\end{equation}
$O_i$ is the theoretical fair throughput for agent $i$; in our algorithm, it is defined as
\begin{equation*}
O_i=\frac{(\text{Maximum data rate})}{(\text{Total spectrum users})}, \ \ \forall i\in[1,N]
\end{equation*}
It is worth to point out that our Dec-POMDP model does not possess an explicit objective state, that is, there is not a state which terminates the mission of all agents when some agents have arrived at the state. In contrast, our model is infinite-horizon, which means theoretically the agent-model interaction will never stop (definitely the agents will stop at some point in practical learning process).

\section{Nonparametric Bayesian Policy Learning}
\label{sec: method}

To accommodate action selection in infinite horizon Dec-POMDP model, we adopt finite state controller for policy representation and utilize Bayesian inference to estimate the parameters of the policy. In this section we introduce the structure of finite state controller and our nonparametric Bayesian learning method.

\subsection{Policy Representation}

\begin{figure}
\centerline{\includegraphics{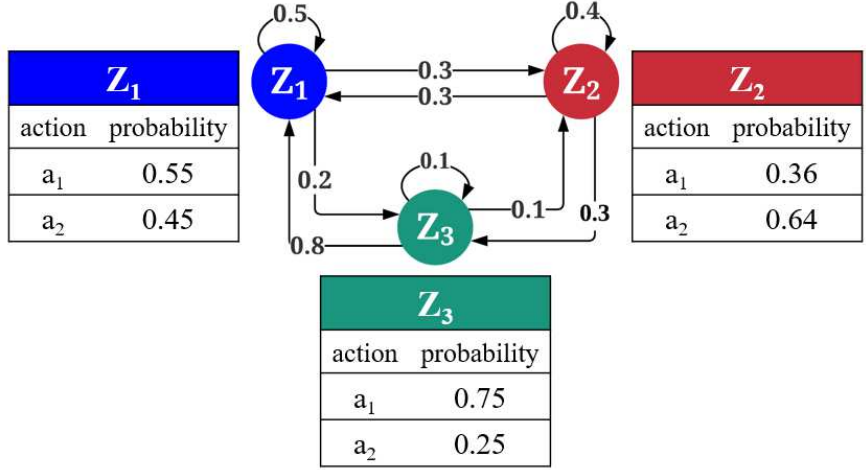}}
	\caption[FSC Policy]{Diagram for a simple FSC policy representation with $|\mathcal{Z}|=3$ and $|\mathcal{A}|=2$. Left: each arch shows the transition probability from one node to another; Right: action probability at each node.}
	\label{fig:FSC}
\end{figure}

Finite State Controller (FSC) is an appropriate policy representation for infinite-horizon stochastic Dec-POMDP models \cite{HansenFSC,AmatoFSC} when the action, observation, and reward space is discrete. It is subsumed a special case of the regionalized policy representation (RPR) \cite{li09multi} when each belief region concentrate to one node. In \cite{li09multi,liu_thesis}, each node in the FSC policy is referred to a decision state or local belief state and treated as latent variables, and integrated out to yield a policy that directly mapping past history of actions and observations to probability distribution over current actions, thus estimation of the true states can be omitted. \ref{fig:FSC} illustrates a simple example of FSC policy with 3 nodes and 2 actions at each node. The FSC policy representation for agent $n$ can be described by a tuple $\langle\mathcal{A}_n,\mathcal{O}_n,\mathcal{Z}_n,\eta_n,\omega_n,\pi_n\rangle$. $\mathcal{A}_n$ and $\mathcal{O}_n$ have been defined in \ref{sec: decPOMDP}; $\mathcal{Z}_n$ is a finite set of nodes; $\eta_n$ is node probability distribution at $t=0$. $\omega_n: \mathcal{Z}_n\times\mathcal{A}_n\times\mathcal{O}_n\rightarrow [0,1]$ is the node transition probability, mapping from node, action, and observation sets to node set, which indicates how the agent will traverse the nodes after an action is performed and an observation is received. $\pi_n$ represents the action selection probability at each node. Each node serves as sufficient statistics of histories of past actions and observations, saving memory space by removing the need of storing histories. Thus FSC is efficient in operating on small devices.
FSC policy representation is suitable for our problem since it can stay simple and compact even for large problem space. Since our problem does not have an explicit end point, a map-like policy representation is not a proper choice. Even though the observation or reward space is enormous, generally there is only a relatively small part of it assigned positive rewards, which is desired for the agent. The cyclic graph of FSC policy captures these necessary parts of our infinite-horizon POMDP model and yields a (ideally) concise framework for the optimal policy, which makes FSC policy popular in various reinforcement learning problems.

\subsection{Nonparametric Policy Prior}

One of the main problems in learning the FSC policies for decentralized agents is determining the sizes of the FSC policies. As we have emphasized, our coexistence problem is dynamic and non-cooperative. With decentralized learning, the local action and observation sets possessed by each agent differ, causing the number of nodes and transition between nodes in different policies deviate from each other. Parametric models impose strong assumption over policy structures, yielding fixed-size policy, which is not applicable to decentralized models. Bounding the space of policy representation may force the learning process to sub-optimal results. In contrast, nonparametric model treats the FSC size as extra variable, which enables it to accommodate unbounded variety of nodes sets and transition probabilities, allowing each agent to optimize its own policy individually. 
\begin{defn}
Providing the FSC policy representation described above, the stick-breaking process is utilized to generate the prior distributions for node transition probabilities $\omega_n$, and Dirichlet distribution is adopted for prior distribution over actions $\pi_n$ at each node. Gamma distribution is placed over $\alpha$ in beta distribution for stick-breaking construction as hierarchical prior \textup{\cite{LiuSBPR}}:
\[
\begin{aligned}
    & \eta_n^1=u_n^1,\ \eta_n^i=u_n^i\prod_{m=1}^{i-1}\left(1-u_n^m\right) \\
    & \omega_{n,a,o}^{i,1}=V_{n,a,o}^{i,1},\  \omega_{n,a,o}^{i,1:j}=V_{n,a,o}^{i,1:j}\prod_{m=1}^{j-1}\left(1-V_{n,a,o}^{i,m}\right) \\
    & u_n^{1:\infty}\sim\textup{Beta}(1,\rho_{n}), \ \ \rho_{1:N}\sim\textup{Gamma}(e,f) \\ 
    & V_{n,a,o}^{i,1:\infty}\sim\textup{Beta}(1,\alpha_{n,a,o}^{i}), \ \ \alpha_{n,a,o}^{1:\infty}\sim\textup{Gamma}(c_{n,a,o},d_{n,a,o}) \\
    & \pi_{n,i}^{1:|\mathcal{A}_n|}\sim\textup{Dirichlet}\left(\theta_{n,i}^{1:|\mathcal{A}_n|}\right)
\end{aligned}
\]
for node indices $i,j=1,...,\infty$
\label{def: SBPR}
\end{defn}
Hyper-parameters $(c, d, e, f, \theta)$ determine the distributions of $\eta$, $\omega$, and $\pi$. $|\cdot|$ represents the cardinality of a set. For notational elegance, we utilize the same abbreviation in \cite{LiuSBPR}. Let consecutive sequence $(i,i+1,...,j)$ reduce to $i:j$, so $(\omega_{n,a,o}^{i,1},...,\omega_{n,a,o}^{i,j})=\omega_{n,a,o}^{i,1:j}$ represents the node transition probabilities from node $i$ to nodes $1,...,j$ for agent $n$, after performing action $a\in\mathcal{A}_n$ and observing $o\in\mathcal{O}_n$. Similarly, $\left(\pi_{n,i}^{1},...,\pi_{n,i}^{|\mathcal{A}_n|}\right)=\pi_{n,i}^{1:|\mathcal{A}_n|}$ means the probabilities of selecting actions $a_1,...,a_{|\mathcal{A}_n|}$ for agent $n$ at node $i$.

\subsection{Global Empirical Value Function}
\begin{figure}
\centerline{\includegraphics[width=1\linewidth]{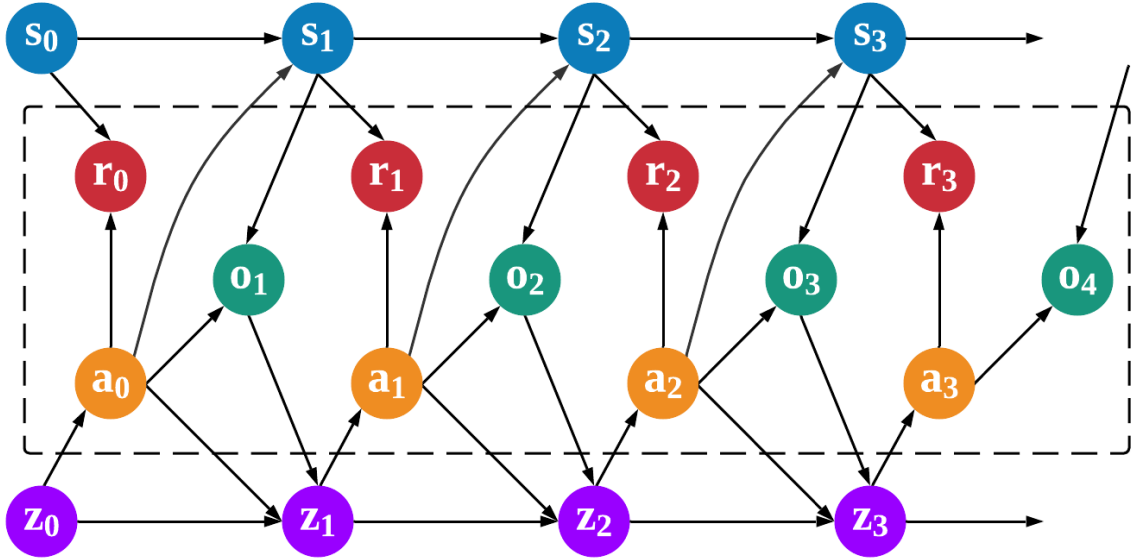}}
	\caption[DBN Expression for POMDP Model]{The dynamic Bayes network graph for an infinite-horizon POMDP model with a FSC policy for one agent, only the components in dash-line box are visible to the agent when learning policy.}
	\label{fig:POMDP}
\end{figure}
\begin{figure}
\centerline{\includegraphics[width=0.9\linewidth]{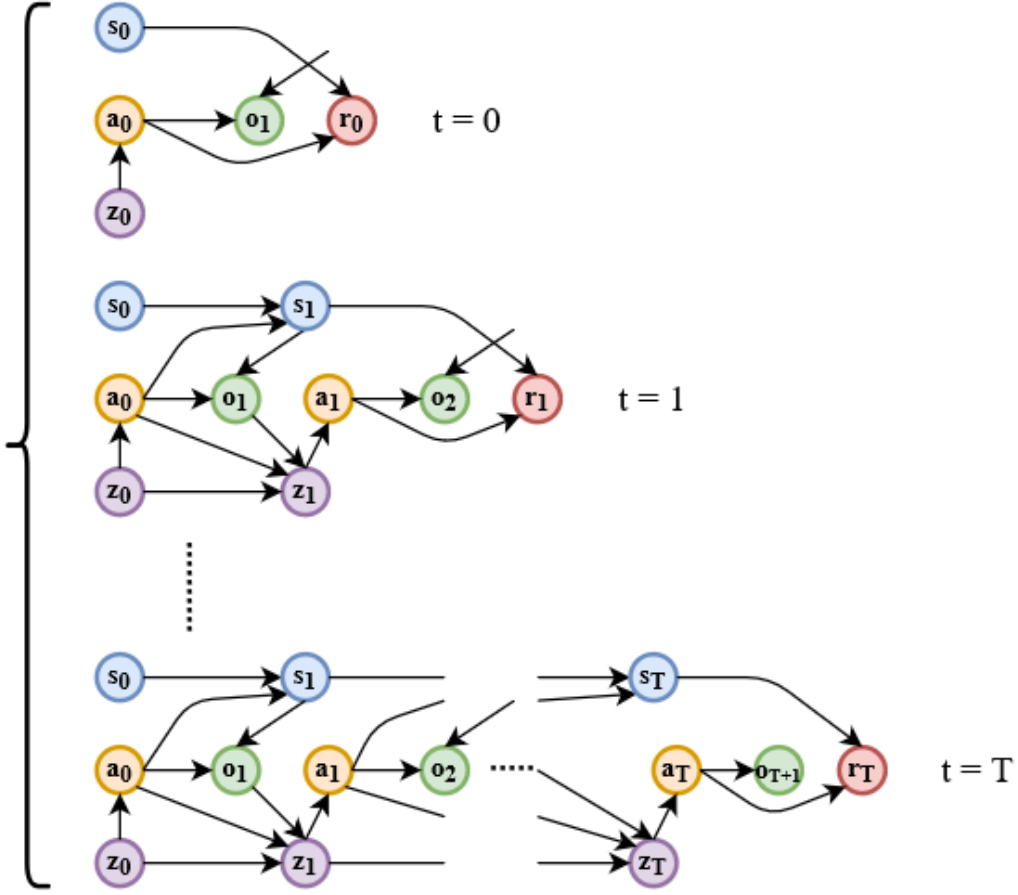}}
	\caption[Mixture of DBNs]{Decomposing the DBN of our POMDP model in \ref{fig:POMDP} into mixture of sub-networks, where reward at each step only emits at the end of each DBN.}
	\label{fig:DBN}
\end{figure}

In general, the objective of reinforcement learning is to maximize the value function. In order to adopt Bayesian approach, the value function is translated into likelihood to exhibit the value of collected data given policy \cite{toussaintDBN}. An Dec-POMDP can be formulated as one single Dynamic Bayes Network (DBN) with a binary reward variable $R$ at each time step. However, this DBN can be decomposed into an infinite mixture of DBNs \cite{kumarDBN}, where reward only emerges at the end of each DBN. \ref{fig:POMDP} illustrates the Bayes network representation of one agent for our Dec-POMDP model including the nodes in FSC policy, where arcs exhibit the dependency between variables; variables in dash-line box are visible to the learning agent. \ref{fig:DBN} represents the result of decomposing the Bayes network in \ref{fig:POMDP} into mixture of sub-networks. There is only one unique DBN for each time length $T=t$. Denote $r_T(\Theta)$ as immediate reward received by following policy $\Theta$ in DBN of length $T$, the value $\hat{r}_T(\Theta)$ obtained by by normalizing $r_T(\Theta)$ into range $[0,1]$ is proportional to the likelihood $p(R=1|T,\Theta)$,
\begin{equation}
    \hat{r}_T(\Theta)=\frac{r_T(\Theta)-R_{\text{min}}}{R_{\text{max}}-R_{\text{min}}}\propto p(R=1|T,\Theta)
\end{equation}
This implies maximizing the likelihood $p(R=1|T,\Theta)$ is equivalent to maximizing the reward. After imposing a geometric distribution with parameter equal to the discount factor $\gamma$ over the mixture of DBNs, the joint likelihood $p(R=1|\Theta)$ is obtained by marginalizing $T$,
\begin{equation}
\begin{aligned}
    p(R=1|\Theta) &=\sum_{t=0}^{T}p(t)p(R=1|t,\Theta) \\
    &=\sum_{t=0}^{T}(1-\gamma)\gamma^t p(R=1|t,\Theta) \\
    &=\sum_{t=0}^{T}(1-\gamma)\gamma^t \frac{r_t(\Theta)-R_{\text{min}}}{R_{\text{max}}-R_{\text{min}}} \\
    &=\frac{1-\gamma}{R_{\text{max}}-R_{\text{min}}}\left[\sum_{t=0}^{T}\gamma^t r_t(\Theta)-\sum_{t=0}^{T}\gamma^tR_{\text{min}}\right] \\
    &=\frac{1-\gamma}{R_{\text{max}}-R_{\text{min}}}\hat{V}(\Theta),
\end{aligned}
\label{equ: likelihood}
\end{equation}
where $\hat{V}(\Theta)$ is the shifted value function given policy $\Theta$. So maximizing this likelihood amounts to maximizing the value of the Dec-POMDP given policy $\Theta$. In \cite{li09multi} Li {\it et al.} proposed an empirical value function $\hat{V}(\mathcal{D}^K;\Theta)$ to acquire the value of desired policy $\Theta$ with $K$ trajectories. 
\begin{defn}
The $k$-th history for agent $n$ from time $0$ to $t$ is defined as the sequence $(a_{n,0}^k,...,a_{n,t-1}^k;o_{n,1}^k,...,o_{n,t}^k)=(a_{n,0:t-1}^k,o_{n,1:t}^k)=h_{n,t}^k$, and the $k$-th trajectory $\mathcal{D}^k$ with length $T_k$ is the sequence $(\vec{a}_0^k,r_0^k,\vec{o}_1^k,...,\vec{o}_{T_k}^k,\vec{a}_{T_k}^k,r_{T_k}^k)$. The value $\hat{V}(\mathcal{D}^K;\Theta)$ is the expected value of discount sum of rewards with respect to reweighted policy:
\[
\begin{aligned}
    & \hat{V}(\mathcal{D}^K;\Theta) \\
    &=\textup{E}_{\Theta}\left[\sum_{k=1}^{K}\sum_{t=0}^{T_k}\gamma^t\frac{(r_t^k-R_{\textup{min}})}{K}\right] \\
    &=\sum_{k=1}^{K}\sum_{t=0}^{T_k}\frac{\prod_{\tau=0}^{t}\prod_{n=1}^{N}p(a_{n,\tau}^k|h_{n,\tau}^k,\Theta)}{\prod_{\tau=0}^{t}\prod_{n=1}^{N}p(a_{n,\tau}^k|h_{n,\tau}^k,\Pi)}\gamma^t\frac{(r_t^k-R_{\textup{min}})}{K}
\end{aligned}
\]
\label{def: emp_likelihood}
\end{defn}
$\prod_{\tau=0}^{t}\prod_{n=1}^{N}p(a_{n,\tau}^k|h_{n,\tau}^k,\Theta)$ can be substituted with $p(\vec{a}_{0:t}^{k},\vec{z}_{0:t}^{k}|\vec{o}_{1:t}^{k},\Theta)$ (proof in \ref{ch: em_value}). $\Theta$ is reweighted by the behavior policy $\Pi$ which is utilized for collecting trajectories. By law of large number, $\hat{V}(\mathcal{D}^K;\Theta)$ approximates $\hat{V}(\Theta)$ as $K$ approaches infinity. \ref{def: emp_likelihood} enables us to utilize existing trajectories to compute the likelihood instead of collecting them ourselves. Combining equation \ref{equ: likelihood} and \ref{def: emp_likelihood}, the likelihood is connected to the empirical value function,
\begin{equation*}
    p(\mathcal{D}^K|\Theta)\propto p(R=1|\Theta)\propto \hat{V}(\mathcal{D}^K;\Theta)
\end{equation*}
\subsection{Variational Inference for Posterior Approximation}

Providing prior distributions and likelihood function, the objective is to infer the posterior distribution $p(\Theta|\mathcal{D}^K)$. By \ref{equ: ELBO} we can derive the expectation term for joint likelihood
\begin{equation}
\medmuskip=1mu
\thinmuskip=1mu
\thickmuskip=1mu
\begin{aligned}
&\text{E}_q\left[\ln \hat{V}(\mathcal{D}^K;\Theta)p(\Theta)p(\rho)p(\alpha)\right]\\
&=\text{E}_q\left[\ln \sum_{k=1}^{K}\sum_{t=0}^{T_k}\frac{\prod_{\tau=0}^{t}\prod_{n=1}^{N}p(a_{n,\tau}^k|h_{n,\tau}^k,\Theta)}{\prod_{\tau=0}^{t}\prod_{n=1}^{N}p(a_{n,\tau}^k|h_{n,\tau}^k,\Pi)}\gamma^t\frac{(r_t^k-R_{\textup{min}})}{K}p(\Theta)p(\rho)p(\alpha)\right] \\
&=\text{E}_q\left[\ln \sum_{k=1}^{K}\frac{1}{K}\sum_{t=0}^{T_k}\Tilde{r}_t^k p(\vec{a}_{0:t}^{k}|\vec{o}_{1:t}^{k},\Theta)p(\Theta)p(\rho)p(\alpha)\right]\\
&=\text{E}_q\left[\sum_{k=1}^{K}\frac{1}{K}\sum_{t=0}^{T_k}\sum_{\vec{z}_{0:t}^k=1}^{|Z|}\ln \left[\Tilde{r}_t^k p(\vec{a}_{0:t}^{k},\vec{z}_{0:t}^{k}|\vec{o}_{1:t}^{k},\Theta)p(\Theta)p(\rho)p(\alpha)\right]\right]\\
&=\sum_{k=1}^{K}\frac{1}{K}\sum_{t=0}^{T_k}\sum_{\vec{z}_{0:t}^k=1}^{|Z|}\int q(\Theta)q(\rho)q(\alpha)q(\vec{z}_{0:t}^k)\ln \Tilde{r}_t^k p(\vec{a}_{0:t}^{k},\vec{z}_{0:t}^{k}|\vec{o}_{1:t}^{k},\Theta)d\Theta d\rho d\alpha \\
&+\sum_{k=1}^{K}\frac{1}{K}\sum_{t=0}^{T_k}\sum_{\vec{z}_{0:t}^k=1}^{|Z|}\text{E}_q\left[\ln p(\Theta)+\ln p(\rho)+\ln p(\alpha)\right] \\
&=\sum_{k=1}^{K}\frac{1}{K}\sum_{t=0}^{T_k}\sum_{\vec{z}_{0:t}^k=1}^{|Z|}\int q(\Theta)q(\vec{z}_{0:t}^k)\ln \Tilde{r}_t^k p(\vec{a}_{0:t}^{k},\vec{z}_{0:t}^{k}|\vec{o}_{1:t}^{k},\Theta)d\Theta \\
&+\text{E}_q\left[\ln p(\Theta)\right]+\text{E}_q\left[\ln p(\rho)\right]+\text{E}_q\left[\ln p(\alpha)\right] \\
&=\text{E}_{q(\Theta,z)}\left[\ln \Tilde{r}_t^k p(\vec{a}_{0:t}^{k},\vec{z}_{0:t}^{k}|\vec{o}_{1:t}^{k},\Theta)\right]+\text{E}_{q(\Theta,\rho,\alpha)}\left[\ln p(\Theta)\right]+\text{E}_{q(\rho)}\left[\ln p(\rho)\right] \\
&+\text{E}_{q(\alpha)}\left[\ln p(\alpha)\right]
\end{aligned}
\label{equ: ELBO_p}
\end{equation}
where $\Tilde{r}_t^k=\gamma^t\frac{r_t^k-R_{\text{min}}}{\prod_{n=1}^{N}p(a_{n,0:t}^k|o_{n,1:t}^k,\Pi)}$. $\Theta$ denotes the policy variables $(u, V, \pi)$. The probability of node transition history $p(z_{n,0:t}^k|a_{n,1:t}^k,o_{n,1:t}^k,\Theta)$ also needs to be inferred since $(\eta, \omega, \pi)$ depend on $z$. Applying mean-field approximation, the expectation over $q$ distribution can be derived
\begin{equation}
    \begin{aligned}
        &\text{E}_q\left[\ln q(\Theta, \rho,\alpha)q(\vec{z}_{0:t}^k)\right]\\
        &=\text{E}_q\left[\ln q(\Theta)q(\rho)q(\alpha)q(\vec{z}_{0:t}^k)\right]\\
        &=\text{E}_q\left[\ln q(\Theta)\right]+\text{E}_q\left[\ln q(\rho)\right]+\text{E}_q\left[\ln q(\alpha)\right]+\text{E}_q\left[\ln q(\vec{z}_{0:t}^k)\right]\\
        &=\text{E}_{q(\Theta)}\left[\ln q(\Theta)\right]+\text{E}_{q(\rho)}\left[\ln q(\rho)\right]+\text{E}_{q(\alpha)}\left[\ln q(\alpha)\right]+\text{E}_{q(z)}\left[\ln \prod_{n=1}^{N}q(z_{n,0:t}^k)\right]
    \end{aligned}
\label{equ: ELBO_q}
\end{equation}
Combining \ref{equ: ELBO_p} and \ref{equ: ELBO_q}, we obtain the $\textup{ELBO}(q)$ as
\begin{equation}
    \textup{ELBO}(q)=\text{E}_q\left[\ln \hat{V}(\mathcal{D}^K;\Theta)p(\Theta)p(\rho)p(\alpha)\right]-\text{E}_q\left[\ln q(\Theta,\rho,\alpha)q(\vec{z}_{0:t}^k)\right]
\label{equ: ELBO_final}
\end{equation}
Mean-field assumption is imposed over the joint variational $q(\Theta)$ to divide it into the product of marginal $q(u)q(V)q(\pi)$ conditional on their corresponding parameters. Since the likelihood is assumed as discrete distribution, the Dirichlet distribution and Dirichlet process we place for policy priors are conjugate prior; thus the true posterior distributions for $(u,V,\pi,\rho,\alpha)$ are reasonably assumed to belong to the same family of their corresponding prior distributions. The variational distributions $q$ for posterior approximation are defined as follows:
\begin{equation}
    \begin{aligned}
    &q(z_{n,0:t}^{k})=\Tilde{\nu}_t^k p(z_{n,0:t}^k|a_{n,0:t}^k,o_{n,1:t}^k,\Tilde{\Theta}), \  \forall (n,k,t) \text{ indices} \\
    &q(u_n^i)=\text{Beta}(\delta_n^i,\mu_n^i), \, \forall (n,i) \text{ indices} \\
    &q(V_{n,a,o}^{i,j})=\text{Beta}(\sigma_{n,a,o}^{i,j},\lambda_{n,a,o}^{i,j}), \  \forall (n,a,o,i,j) \text{ indices} \\
    &q(\rho_n)=\text{Gamma}(g_n,h_n), \  \forall n \text{ indices} \\
    &q(\alpha_{n,a,o}^{i})=\text{Gamma}(a_{n,a,o}^{i},b_{n,a,o}^{i}), \  \forall (n,a,o,i) \text{ indices} \\
    &q(\pi_{n,i})=\text{Dirichlet}\left(\phi_{n,i}^{1},...,\phi_{n,i}^{|\mathcal{A}_n|}\right), \  \forall (n,i) \text{ indices} \\
    &\Tilde{\nu}_t^k =\gamma^t (r_t^k-R_{\text{min}}) \frac{\prod_{n=1}^{N}p(a_{n,0:t}^{k}|o_{n,1:t}^{k}, \Tilde{\Theta})}{\prod_{n=1}^{N}p(a_{n,0:t}^{k}|o_{n,1:t}^{k}, \Pi)\hat{V}(\mathcal{D}^K;\Tilde{\Theta})} 
\end{aligned}
\label{equ: q_dist}
\end{equation}
$\Tilde{\Theta}=(\Tilde{\eta},\Tilde{\pi},\Tilde{\omega})$ is the point estimate of optimal policy parameters from previous iteration of variation inference. It is worth to note that each node transition probability $q(z_{n,t}^{k})$ is a multinomial distribution. By placing Dirichlet process prior over it, we can approximate the posterior with mean-field variational distribution $q_{n,t}^k(z_{n,0:t}^{k})$. For simplicity, expectation maximization approach is adopted for $q_{n,t}^k(z_{n,0:t}^{k})$ \cite{li09multi}. By taking derivative on $\text{ELBO}(q)$ with respect to each $q$ distribution and set as zero while keeping all others fixed, the Coordinate Ascent VI (CAVI) is adopted to obtain each optimal $q^*$ distribution
\begin{theorem}
With conjugate prior and mean-field approximation, the derivation of each variational distribution can be reduce to the parameter computation for each $q$ in \textup{\ref{equ: q_dist}}:
\[
\medmuskip=1mu
\thinmuskip=1mu
\thickmuskip=1mu
\begin{aligned}
    &\delta_n^i=1+\sum_{k=1}^{K}\frac{1}{K}\sum_{t=0}^{T_k}q_{n,t}^k(z_{n,0}^{k}=i) \\
    &\mu_n^i=\frac{g_n}{h_n}+\sum_{k=1}^{K}\frac{1}{K}\sum_{t=0}^{T_k}\sum_{m=i+1}^{|\mathcal{Z}_n|}q_{n,t}^k(z_{n,0}^{k}=m) \\
    &\phi_{n,i}^{a}=\theta_{n,i}^{a}+\sum_{k=1}^{K}\frac{1}{K}\sum_{t=0}^{T_k}\sum_{\tau=0}^{t}q_{n,t}^k(z_{n,\tau}^{k}=i)\mathbb{I}(a_{n,\tau}^{k}=a) \\
    &\sigma_{n,a,o}^{i,j}=1+\sum_{k=1}^{K}\frac{1}{K}\sum_{t=0}^{T_k}\sum_{\tau=1}^{t}q_{n,t}^k(z_{n,\tau-1}^{k}=i,z_{n,\tau}^{k}=j)\mathbb{I}(a_{n,\tau-1}^{k}=a,o_{n,\tau}^{k}=o) \\
    &\lambda_{n,a,o}^{i,j}=\frac{a_{n,a,o}^{i}}{b_{n,a,o}^{i}}+\sum_{k=1}^{K}\frac{1}{K}\sum_{t=0}^{T_k}\sum_{\tau=1}^{t}\sum_{m=j+1}^{|\mathcal{Z}_n|}q_{n,t}^k(z_{n,\tau-1}^{k}=i,z_{n,\tau}^{k}=m)\mathbb{I}(a_{n,\tau-1}^{k}=a,o_{n,\tau}^{k}=o) \\
    &g_n=e+|\mathcal{Z}_n|, \ \ h_n=f-\sum_{i=1}^{|\mathcal{Z}_n|}\left[\Psi(\mu_{n}^{i})-\Psi(\delta_{n}^{i}+\mu_{n}^{i})\right] \\
    &a_{n,a,o}^{i}=c_{n,a,o}+|\mathcal{Z}_n|, \ \ b_{n,a,o}^{i}=d_{n,a,o}-\sum_{j=1}^{|\mathcal{Z}_n|}\left[\Psi(\lambda_{n,a,o}^{i,j})-\Psi(\sigma_{n,a,o}^{i,j}+\lambda_{n,a,o}^{i,j})\right]
\end{aligned}
\]
where
\[
\begin{aligned}
    & q_{n,t}^k(z_{n,\tau}^{k}=i)=\Tilde{\nu}_t^k p(z_{n,\tau}^{k}=i\vert a_{n,0:t}^k,o_{n,1:t}^k,\Tilde{\Theta}) \\
    & q_{n,t}^k(z_{n,\tau-1}^{k}=i,z_{n,\tau}^{k}=j)=\Tilde{\nu}_t^k p(z_{n,\tau-1}^{k}=i,z_{n,\tau}^{k}=j\vert a_{n,0:t}^k,o_{n,1:t}^k,\Tilde{\Theta})
\end{aligned}
\]
are marginal distributions of $q_{n,t}^k(z_{n,0:t}^{k})$ for $\tau=0,...,t$.
\label{thm: q_update}
\end{theorem}
$\Psi(\cdot)$ is the digamma function. The detail of \ref{thm: q_update} is presented in \ref{ch: VI_update}. Each Bayesian learning iteration for our Dec-POMDP model is exhibited in the following

\begin{algorithm}
\SetKwInOut{Input}{Input}\SetKwInOut{Output}{Output}\SetKwInOut{Initialize}{Initialize}\SetKw{Return}{Return}
\SetKwIF{If}{ElseIf}{Else}{if}{}{}{}{end}
\SetAlgoLined
\Input{$p(\Theta_n)$, $p(\rho_n)$, $p(\alpha_n)$, trajectories $\mathcal{D}^k$, $k=1,...,K$}
\Initialize{initial $\textsc{ELBO}_0(q)$}
 \For{ $\textup{Iter}=1,...,\textup{max}$}{
  Update $\Tilde{\Theta}_n=(\Tilde{\eta}_n,\Tilde{\omega}_n,\Tilde{\pi}_n)$ for $n=1,...,N$ \\
  Compute each marginal $q_{n,t}^k(z_{n,\tau}^k)$ for $\tau=0,...,T_k$\\
  Compute each $q^*(\Theta_n)$, $q^*(\rho_n)$, and $q^*(\alpha_n)$ according to \ref{thm: q_update}\\
  Compute $\textsc{ELBO}_{\textup{Iter}}(q)$ by \ref{equ: ELBO_final} \\
  $\Delta\textsc{LB}(q)=\vert(\textsc{ELBO}_{\textup{Iter}}(q)-\textsc{ELBO}_{\textup{Iter}-1}(q))/\textsc{ELBO}_{\textup{Iter}-1}(q)\vert$ \\
  \If{$\Delta\textsc{LB}(q) < 10^{-5}$}{
   break\;
   }
 }
\Output{variational distributions $q^*(\Theta_n)$, $q^*(\rho_n)$, and $q^*(\alpha_n)$ for $n=1,...,N$}
\caption{CAVI for Bayesian Reinforcement Learning}
\label{alg: learning}
\end{algorithm}
\chapter{SIMULATIONS}
\label{chap: simu}
\acresetall
In this chapter we detail the system setup for performance evaluation of our Bayesian reinforcement learning algorithm and demonstrate the simulation results along with discussions.
%
\begin{table}[htbp]
\begin{center}
\begin{tabular}{|l|l|}
\hline
\textbf{Parameter Name} & \textbf{Value} \\
\hline \hline 
Number of LTE eNB & 2 \\ 
Number of Wi-Fi AP & 2 \\ 
Number of channel & 1 \\
Channel bandwidth & 20~MHz \\
DIFS duration & 34~$\mu$s \\
Wi-Fi back-off slot & 9~$\mu$s \\
ICCA duration & 43~$\mu$s \\
ECCA slot & 9~$\mu$s \\
Contention window & 15,31,63,127,255,511,1023 \\
LTE sub-frames per transmission & 3,6,8,10~ms \\
Wi-Fi packets per transmission & 1 \\
size of Wi-Fi packet & 15000~bytes \\
Transmission rate & 30~Mbps \\
Discount factor $\gamma$ & 0.9 \\
\hline
\end{tabular}
\end{center}
\caption{Pre-Defined Parameters}
\label{tab: simu_param}
\vspace{-5mm}
\end{table}

Due to limited computing resource, we only performed our simulation with small data set to obtain the results. \ref{tab: simu_param} lists the parameters for establishing our simulation environment \cite{3gpp.36.213}. For simplicity, the sets of available contention windows, are identical for both LTE and Wi-Fi agents. Our scenario simulates the spectrum sharing in time domain, which mean only one channel can be accessed by wireless agents; frequency domain multiplexing is beyond our scope. In \cite{3gpp.36.213}, the maximum channel occupation time and contention window for LTE agents differ with spectrum access priorities. In our scenario, the channel occupation time a LTE agent can utilize depends on the contention window it selects in consideration of fair coexistence with Wi-Fi agents. For instance, if a LTE agent selects window size 15 for its back-off sensing, then it can occupy the channel for 3ms after the channel sensing is finished. This occupation time is 6ms for window sizes $\{31, 63\}$, 8ms for window sizes $\{127, 255\}$, and 10ms for window sizes $\{511, 1023\}$. Wi-Fi packet is fixed whichever the contention window it selects. Each Wi-Fi packet amounts of 15000 bytes including overhead. During back-off sensing, the agent perform spectrum sensing each 1$\mu$s to judge whether the channel is clear. However, the correctness of judgement is affected by path loss, fading, and shadowing effect between the sensing and transmitting agents. Each active agent has probability $p_e$ to be judged as idle when it is occupying the channel. Each back-off sensing slot is considered as clear if the channel is assessed as busy no more than 5$\mu$s out of 9$\mu$s. A Wi-Fi packet or LTE sub-frame is assumed to be lost if collision happens during its transmission, and one fail sub-frame does not affect other sub-frames in the same LTE transmission. Finally, all wireless agents have infinite amount of data to transfer, which and all wireless nodes access the same channel. For discrete model, rewards and observations are rounded to integer values. 

\section{Performance Evaluation}
\label{sec: perform_eval}
For the optimization of \ref{alg: learning}, the learning of FSC policies for all agents are based on $K=10$ episodes with each episode of length $T=50$. To accelerate the optimization, cross validation was implemented for better initializing the hyper-parameters of the prior distributions in our variational inference. In our simulation, the hyper-parameters are set to $c=e=0.1$, $d=f=100$ in order to pursuit the minimum optimal FSC policy. TO obtain the initial size of the FSC policy for for each agent, all episodes collected are converted into FSC structures by adopting method similar to \cite{AmatoFSCinit}.
\begin{figure}[t]
    \centering
    \begin{subfigure}[b]{0.5\textwidth}
        \centering
        \includegraphics[width=\textwidth]{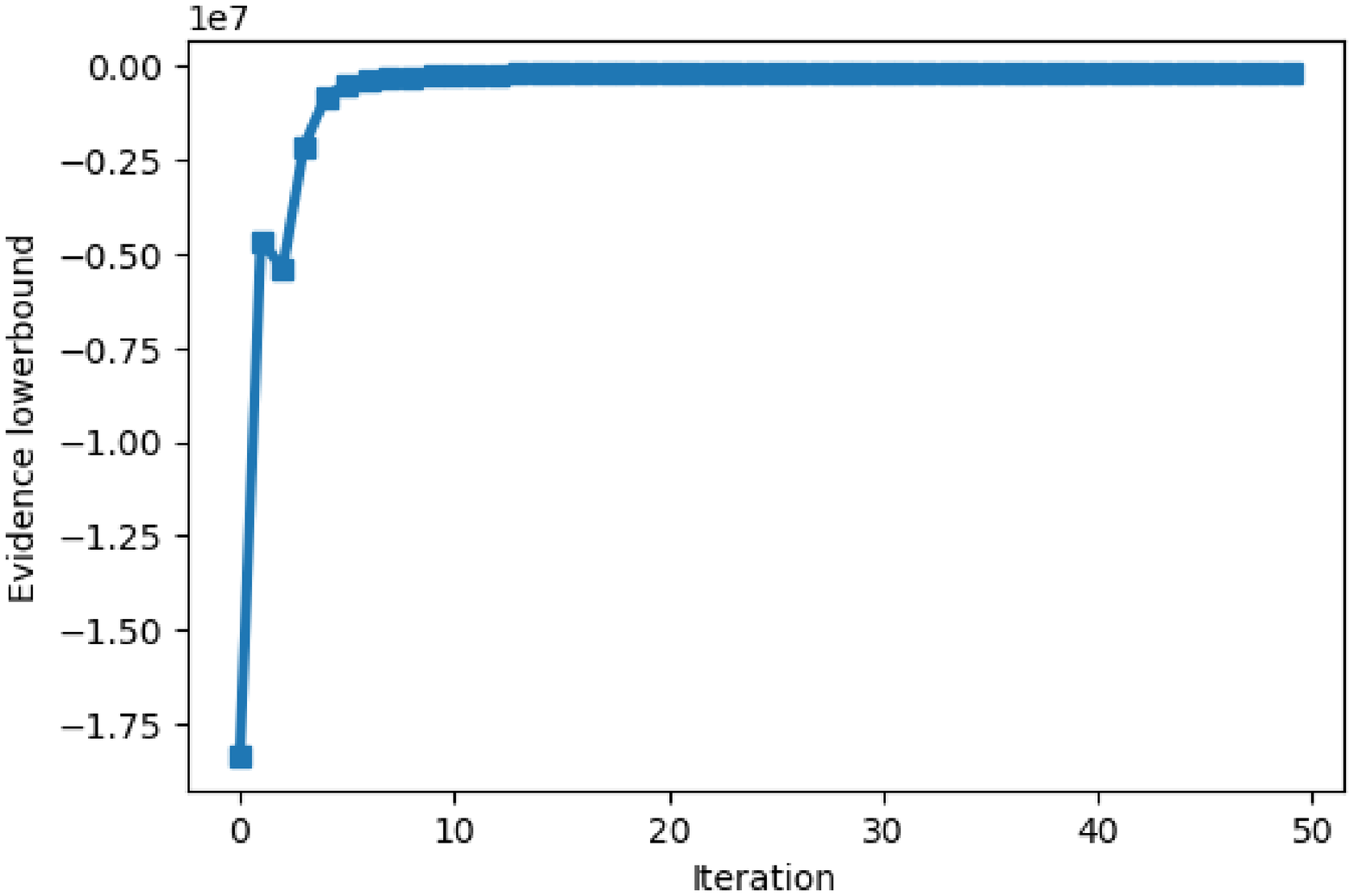}
        \caption[]{{\small ELBO value}}
        \label{fig:ELBO_evolution}
    \end{subfigure}
    \hfill
    \begin{subfigure}[b]{0.477\textwidth}  
        \centering 
        \includegraphics[width=\textwidth]{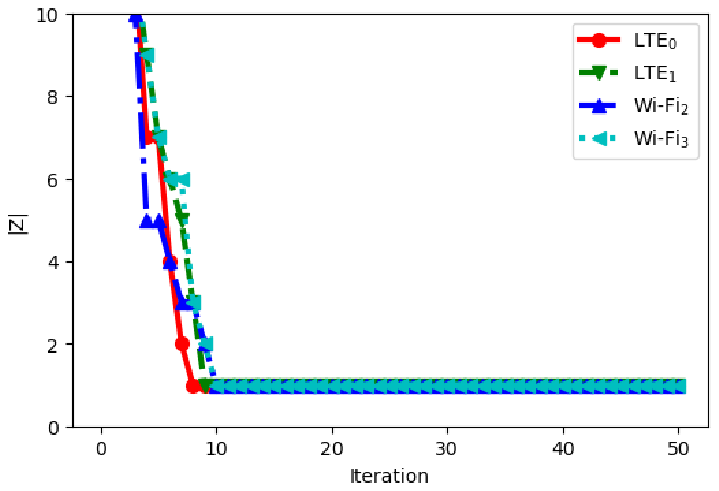}
        \caption[]{{\small Number of FSC nodes}}
        \label{fig:Z_evolution}
    \end{subfigure}
\caption[Evolution of The ELBO Value and Policy Size]{Evolution of the ELBO value and policy size. (a) The convergence of evidence lower bound, (b) The arameter $h$ are fluctuating around a certain level for each agent.}
\label{fig:ELBO_Z_evolution}
\end{figure}
\begin{figure}[t]
\centerline{\includegraphics{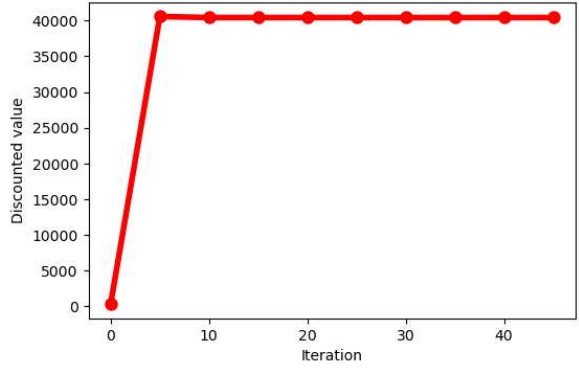}}
	\caption[Discount Value]{Evolution of the discount values with the variational inference.}
	\label{fig:discount_value}
\end{figure}
\begin{figure}[t]
    \centering
    \begin{subfigure}[b]{0.483\textwidth}
        \centering
        \includegraphics[width=\textwidth]{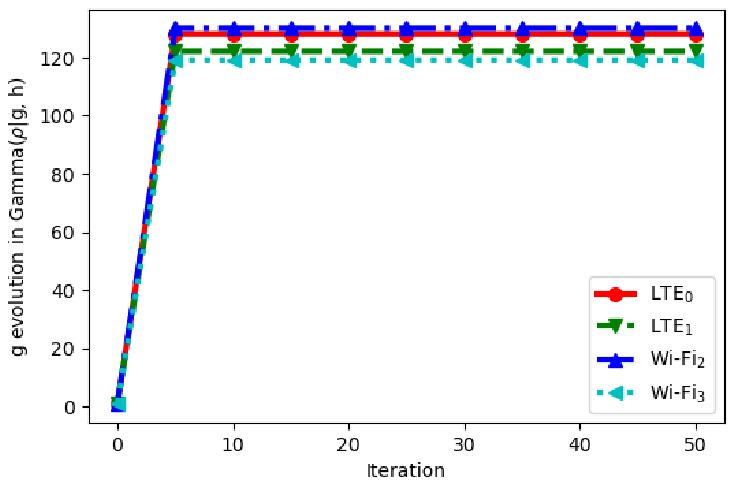}
        \label{fig:g_history}
    \end{subfigure}
    \hfill
    \begin{subfigure}[b]{0.5\textwidth}  
        \centering 
        \includegraphics[width=\textwidth]{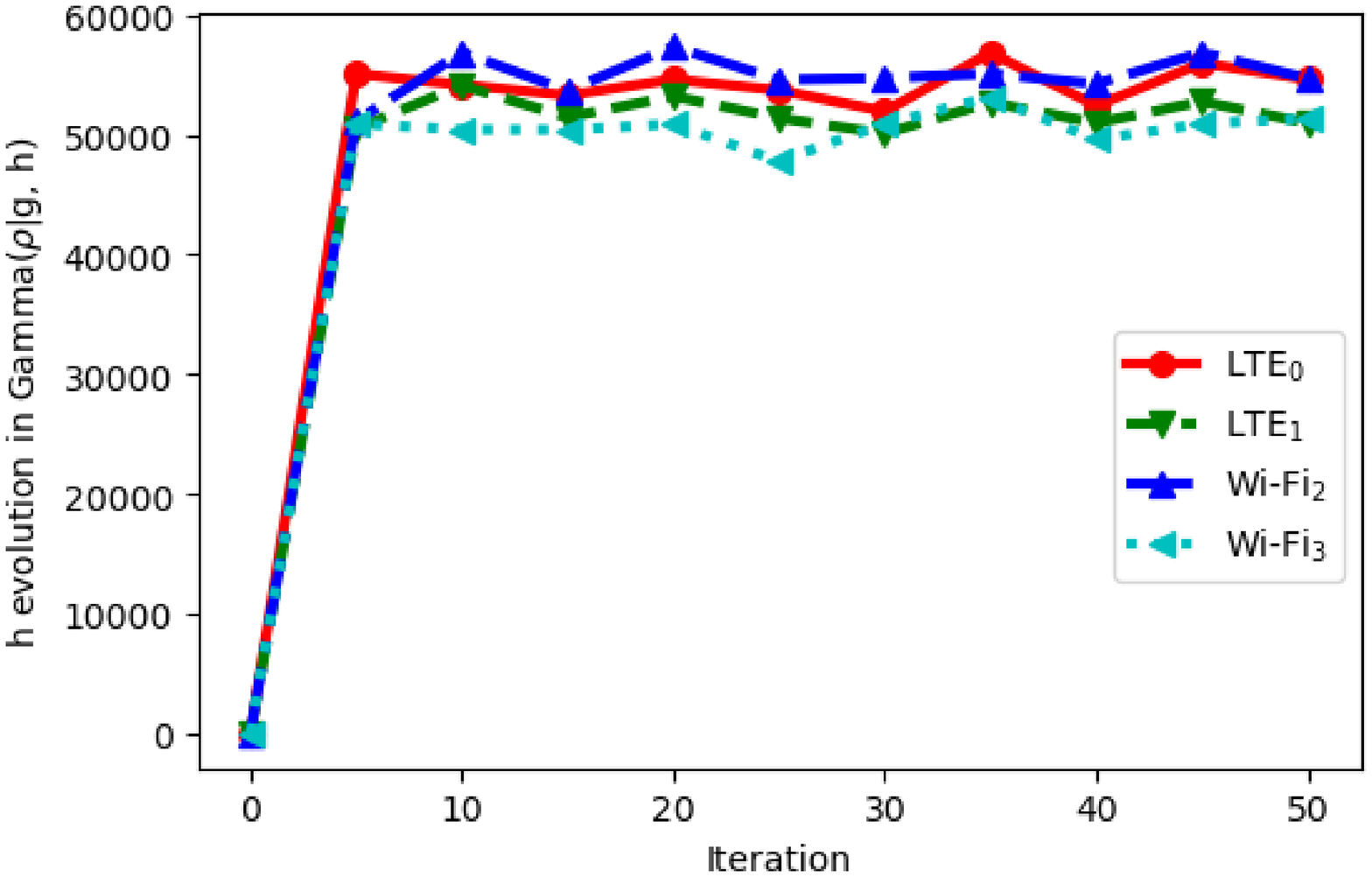}
        \label{fig:h_history}
    \end{subfigure}
\caption[Evolution of The Parameters for $q(\rho|g, h)$]{Evolution of the parameters for $q(\rho|g, h)$. The parameter $g$ keeps constant during the variational inference, while $h$ are fluctuating around a certain level for each agent.}
\label{fig:para_evolution}
\end{figure}
\subsection{Convergence of Variational Inference}
\ref{fig:ELBO_evolution} illustrates the convergence of the evidence lower bound. As iteration was proceeding, the lower bound value was ascending fast to a certain level and the iteration stopped when the value fluctuation satisfied the stop criterion. As lower bound was converging, \ref{fig:Z_evolution} demonstrates how the sizes of the FSC policies for all agents gradually optimized to the minimum values required for maximizing the lower bound. It is interesting to note that with the rich-gets-richer property, the number of nodes for FSC policies all shrinked to $1$ while optimizing the lower bound and discounted value, which means the policies reduced to one-state controller similar to multi-armed bandit. The evolution of the discounted value, which is the sum of discounted rewards for all trajectories, as a function of iteration of the variational inference is exhibited in \ref{fig:discount_value}. Similar to the convergence of lower bound, the ascending discounted value verified the improvement of FSC policies through iterations. The evolution of parameters $(g,h)$ for $q(\rho)$ for each agent is demonstrated in \ref{fig:para_evolution}; as shown in \ref{thm: q_update}, the update of $g$ is a constant for each agent while the value of $h$ fluctuated at different level since it is jointly optimized with other $q$ distributions.

\chapter{CONCLUSIONS}
\label{chap: conclusion}
\acresetall
As wireless technology advances, the coexistence problem in unlicensed spectrum has been an urgent issue waiting for solutions. Many solutions have been proposed, however they all left evident problems yet to be answered. We relaxed assumptions imposed in previous works and exhibited a model close to the real application. Besides that, reinforcement learning is a thriving topic and has been adopted in many fields of applications, including coexistence. Bayesian method over reinforcement learning provides a solution to encode prior knowledge so that the need for large data set is reduced. Nonparametric model over priors allows the learned result to be determined by what the agent has observed without being restricted by the prior setting and simplifies the learning process while sill obtaining excellent result. The combination of empirical value function and variational inference transformed the process of iteratively updating Bellman equation into optimization process, which is superior to conventional reinforcement learning method when the problem model is scaling up.

\section{Vignette of Contributions}
In this work We formulated a real-world spectrum coexistence problem as a Dec-POMDP model and utilized reinforcement learning to explore the optimal channel access policies. An asynchronous model was established for decentralized agents to cooperate for a global interest and a novel cumulative reward function was proposed to incorporate the time dependency of over action and observation history. The Jain's fairness indicator was introduced in reward to balance the spectrum access rights among agents. To adapt to the multiple decentralized learning agents, the Dirichlet process was placed over policy priors to accommodate variable-sized policy representations. This is the first work to consider unbounded model sizes over policy prior in Bayesian reinforcement learning for spectrum coexistence. For posterior inference, arduous sampling methods were replaced by coordinate ascend variational inference, an optimization alternative which turns the distribution approximation into deterministic computation for variational distributions so that scaling up to large problem models was much easier and computationally efficient. We also worked out the computation equations for all variational distributions and demonstrated the ease of implementing them on computer. Simulation results illustrated the efficiency and robustness of such combination; as the evidence lower bound was converging, the value for learned policy was also ascending to an optimal level. The policy size also converged to a lower value, with evolution of parameters of variational distributions stabilized at certain level in accordance with the computation equations.

\section{Future Works}
A never-ending question in reinforcement learning is how to determine the exploration or exploitation during learning process. When performing exploitation, the agent utilizes the optimal result obtained so far while exploration means to probe the potential of higher reward. Generally, exploration should be more encouraged to explore new possibility in the early phase of learning process. As learning process proceeds, the unknown dynamic of the world model becomes less and less, exploitation gradually takes over to finalize the policy. The core is how the curve of exploration rate shapes. A plummeting curve could incur premature learning result while flat curve may fail to converge. In this work the most commonly-utilized $\epsilon$-greedy method is adopted to determine the exploration rate through learning process. In $\epsilon$-greedy method, a parameter $\epsilon\in (0,1)$ is exploited to determine exploration or exploitation when selecting action. In each learning iteration whenever the agent is going to select an action for data collection, a value $u$ is uniformly sampled from interval $[0,1]$, if $u>\epsilon$, the agent performs exploitation and selects action based learned policy so far, otherwise the action will be selected uniformly from all actions available for exploration. To demonstrate the trade-off between exploration and exploitation, two different $\epsilon$-greedy rates are implemented: both start with exploration rate $0.9$ but one ends at value $0.5$ and the other ends at $0.2$. Both learning processes are trained with $40$ iterations. After each learning iteration, the learning result of each $\epsilon$ rate will be evaluated by the mean reward obtained from $20$ episodes with each length of $50$.

In addition to what is mentioned above, there are still outstanding questions yet to be discussed in this work, as well as interesting improvements which can be introduced to advance the learning result. We list some of them here:
\begin{itemize}[leftmargin=*]
\item {\em Dependent nonparametric model for priors:} since our reward function is dependent of past rewards, a dependent prior model which incorporates previous learned result is capable of better utilizing the knowledge the agent has accumulated so far to converge the posterior inference faster.
\item {\em Uneven priorities for agents:} we only considered equal priorities for all agents in this work , however, in real-world application wireless nodes can have different priority levels. If some agents belong to different priority group, weighting factor may be incorporated in the learning process to reflect the changing priorities of different agents.
\item {\em Joint optimization for global and local interests:} in our model there is only one global reward for all agents to optimize, but if each agent can observe its local reward, the efforts contributed to the local and global rewards may need to be balanced with importance weight.
\item {\em Different design of reward function:} the design of reward function implies the ultimate objective of the learning agent, with different performance measurement there can be different design for the reward function.
\item {\em Simulation with large data set:} due to the lack of computing resource, we only simulated our model with small data set. A more complete experiment with more data and large problem model can be performed to demonstrate the robustness of our algorithm.
\end{itemize}

{\singlespace
\addcontentsline{toc}{part}{REFERENCES}
\bibliographystyle{IEEEtran}
\bibliography{Reference.bib}}
\renewcommand{\chaptername}{APPENDIX}
\addtocontents{toc}{APPENDIX \par}
\begingroup
\renewcommand{\clearpage}{}
\onecolumn[\appendix] 
\endgroup
\begin{appendices}
\vspace{-2cm}
\chapter{LIST OF ACRONYMS}
\begin{acronym}
\acro{5G}{fifth generation}
\acro{3GPP}{third generation partnership project}
\acro{IoT}{internet of things}
\acro{LBT}{listen before talk}
\acro{CSMA/CA}{carrier sense multiple access/collision avoidance}
\acro{LTE}{long-term evolution}
\acro{LTE-A}{long-term evolution-advanced}
\acro{LTE-U}{long-term evolution-unlicensed}
\acro{ABS}{almost blank subframe}
\acro{LTE-LAA}{long-term evolution-licensed assisted access}
\acro{GMM}{Gaussian mixture model}
\acro{MCMC}{Markov chain Monte Carlo}
\acro{ML}{maximum likelihood}
\acro{eNB}{evolved node B}
\acro{AP}{access point}
\acro{DIFS}{distributed inter-frame spacing}
\acro{CCA}{clear carrier assessment}
\acro{ICCA}{initial clear carrier assessment}
\acro{ECCA}{extended clear carrier assessment}
\acro{CW}{contention window}
\acro{DP}{Dirichlet process}
\acro{SB}{stick-breaking}
\acro{VI}{variational inference}
\acro{CAVI}{coordinate ascent variational inference}
\acro{DBN}{dynamic Bayes network}
\acro{ELBO}{evidence lower bound}
\acro{RL}{reinforcement learning}
\acro{MDP}{Markov decision process}
\acro{NN}{neural network}
\acro{RPR}{regionalized policy representation}
\acro{POMDP}{partially-observable Markov decision process}
\acro{Dec-POMDP}{decentralized partially-observable Markov decision process}
\end{acronym}
\end{appendices}

%

\begin{appendices}
\chapter{EMPIRICAL VALUE FUNCTION}
\label{ch: em_value}
To prove that $p(a_{0:t}|o_{1:t})=\prod_{\tau=0}^{t}p(a_{\tau}|h_{\tau})=\prod_{\tau=0}^{t}p(a_{\tau}|a_{0:\tau},o_{1:\tau-1})$, we expand
\begin{equation}
\begin{aligned}
    &p(a_{0:t}|o_{1:t}) \\
    &=\sum_{z_0=1}^{\mathcal{Z}|}\cdots\sum_{z_t=1}^{|\mathcal{Z}|}p(a_{0:t},z_{0:t}|o_{1:t})\\
    &=\sum_{z_0=1}^{|\mathcal{Z}|}\cdots\sum_{z_t=1}^{|\mathcal{Z}|}p(z_0)p(a_0|z_0)\prod_{\tau=1}^{t}p(z_{\tau}|z_{\tau-1},a_{\tau-1},o_{\tau})p(a_{\tau}|z_{\tau})
\end{aligned}
\end{equation}
And since observation $o_t$ does not influence action before time $t$,
\begin{equation}
\begin{aligned}
    &p(a_{0:t-1}|o_{1:t}) \\
    &=\sum_{z_0=1}^{|\mathcal{Z}|}\cdots\sum_{z_t=1}^{|\mathcal{Z}|}\sum_{a_t=1}^{|\mathcal{A}|}p(a_t,a_{0:t-1},z_{0:t}|o_{1:t})\\
    &=\sum_{z_0,...,z_{t}=1}^{|\mathcal{Z}|}\sum_{a_t=1}^{|\mathcal{A}|}\left[p(z_0)p(a_0|z_0)\prod_{\tau=1}^{t-1}p(z_{\tau}|z_{\tau-1},a_{\tau-1},o_{\tau})p(a_{\tau}|z_{\tau})\right]\\
    &\times p(z_{t}|z_{t-1},a_{t-1},o_{t})p(a_t|z_t)\\
    &=\sum_{z_0,...,z_{t-1}=1}^{|\mathcal{Z}|}\left[p(z_0)p(a_0|z_0)\prod_{\tau=1}^{t-1}p(z_{\tau}|z_{\tau-1},a_{\tau-1},o_{\tau})p(a_{\tau}|z_{\tau})\right] \\
    &\times\sum_{z_t=1}^{|\mathcal{Z}|}\sum_{a_t=1}^{|\mathcal{A}|}p(z_{t}|z_{t-1},a_{t-1},o_{t})p(a_t|z_t)\\
    &=\sum_{z_0,...,z_{t-1}=1}^{|\mathcal{Z}|}\left[p(z_0)p(a_0|z_0)\prod_{\tau=1}^{t-1}p(z_{\tau}|z_{\tau-1},a_{\tau-1},o_{\tau})p(a_{\tau}|z_{\tau})\right]\\
    &=\sum_{z_0,...,z_{t-1}=1}^{|\mathcal{Z}|}p(a_{0:t-1},z_{0:t-1}|o_{1:t-1})\\
    &=p(a_{0:t-1}|o_{1:t-1})
\end{aligned}
\label{equ: B2}
\end{equation}
Decompose each $p(a_{\tau}|h_{\tau})$ as
\begin{equation}
    p(a_{\tau}|h_{\tau})=p(a_{\tau}|a_{0:\tau-1},o_{1:\tau})=\frac{p(a_{0:\tau}|o_{1:\tau})}{p(a_{0:\tau-1}|o_{1:\tau})}=\frac{p(a_{0:\tau}|o_{1:\tau})}{p(a_{0:\tau-1}|o_{1:\tau-1})}
\label{equ: B3}
\end{equation}
Combine \ref{equ: B2} and \ref{equ: B3}, there is
\begin{equation}
\begin{aligned}
    &\prod_{\tau=0}^{t}p(a_{\tau}|h_{\tau}) \\
    &=p(a_{t}|a_{0:t-1},o_{1:t})p(a_{t-1}|a_{0:t-2},o_{1:t-1})\cdots p(a_1|a_0,o_1)p(a_0)\\
    &=\frac{p(a_{0:t}|o_{1:t})}{p(a_{0:t-1}|o_{1:t-1})}\frac{p(a_{0:t-1}|o_{1:t-1})}{p(a_{0:t-2}|o_{1:t-2})}\cdots\frac{p(a_{0:1}|o_{1})}{p(a_{0})}p(a_0)\\
    &=p(a_{0:t}|o_{1:t})
\end{aligned}
\end{equation}
\end{appendices}
\begin{appendices}
\chapter{COMPUTATION OF VARIATIONAL DISTRIBUTIONS}
\label{ch: VI_update}

Here we provide the proof of \ref{thm: q_update}. From \ref{equ: VIoptformula} the optimal $q$ distribution for each variable can be obtained by taking derivative on $\text{ELBO}(q)$ with respect to the desired $q$ distribution. The $\text{ELBO}(q)$ for our problem has been derived in \ref{equ: ELBO_p} to \ref{equ: ELBO_final}. By taking derivative on \ref{equ: ELBO_final} with respect to each $q\left(z_{n,0:t}^k\right)$, $q(\Theta_n)$, $q(\rho_{n})$, and $q\left(\alpha_{n,a,o}^{i}\right)$ respectively while keeping all others fixed then reorganize it in terms of the distribution form defined in \ref{equ: q_dist}, each optimal $q$ distribution can be computed analytically.

For $q\left(z_{n,0:t}^k\right)$, keep all $q(\Theta_{n})$, $q(\rho_n)$, and $q\left(\alpha_{n,a,o}^{i}\right)$ fixed, the optimal $q^*\left(z_{n,0:t}^k\right)$ is obtained from $\frac{\partial}{\partial q\left(z_{n,t}^k\right)}\text{ELBO}(q)=0$ with constraint
\begin{equation}
    \sum_{k=1}^K\frac{1}{K}\sum_{t=0}^{T_k}\sum_{z_{1:N,0}^k=1}^{|\mathcal{Z}|}\cdots\sum_{z_{1:N,t}^k=1}^{|\mathcal{Z}|}\prod_{n=1}^{N}q\left(z_{n,0:t}^{k}\right)=1, \, \forall (n,k,t) \text{ indices}
\label{equ:q(z)_constraint}
\end{equation}
we have
\begin{equation}
\begin{aligned}
&\frac{\partial}{\partial q\left(z_{n,t}^k\right)}\text{ELBO}(q) \\
&=\sum_{k=1}^{K}\frac{1}{K}\sum_{t=0}^{T_k}\sum_{\Vec{z}_{0}^{k}...\Vec{z}_{t}^{k}=1}^{|\mathcal{Z}|}\int\prod_{i\neq n}q\left(z_{i,0:t}^{k}\right)q(\Theta)\ln\Tilde{r}_{t}^{k}\prod_{n=1}^{N}p\left(a_{n,0:t}^{k},z_{n,0:t}^{k}\vert o_{n,1:t}^{k},\Theta\right)d\Theta \\
&-\sum_{k=1}^{K}\frac{1}{K}\sum_{t=0}^{T_k}\sum_{\Vec{z}_{0}^{k}...\Vec{z}_{t}^{k}=1}^{|\mathcal{Z}|}\prod_{i\neq n}q\left(z_{i,0:t}^{k}\right)\left[\ln\prod_{i=1}^{N}q\left(z_{i,0:t}^{k}\right)\right] \\
&-\sum_{k=1}^{K}\frac{1}{K}\sum_{t=0}^{T_k}\sum_{\Vec{z}_{0}^{k}...\Vec{z}_{t}^{k}=1}^{|\mathcal{Z}|}\prod_{i\neq n}q\left(z_{i,0:t}^{k}\right) \\
&=0
\end{aligned}
\end{equation}
$q(\rho)$ and $q(\alpha)$ integrate out in above equation since they are not directly associated to $z_{n,t}^k$. Remove all terms unrelated to $q(z_{n,t}^k)$ and rearrange terms, we obtain
\begin{equation}
\begin{aligned}
&\sum_{k=1}^{K}\frac{1}{K}\sum_{t=0}^{T_k}\sum_{\Vec{z}_{0}^{k}...\Vec{z}_{t}^{k}=1}^{|\mathcal{Z}|}\int\prod_{i\neq n}q(z_{i,0:t}^{k})q(\Theta)\ln\Tilde{r}_{t}^{k}\prod_{n=1}^{N}p\left(a_{n,0:t}^{k},z_{n,0:t}^{k}\vert o_{n,1:t}^{k},\Theta\right)d\Theta \\
&=\sum_{k=1}^{K}\frac{1}{K}\sum_{t=0}^{T_k}\sum_{\Vec{z}_{0}^{k}...\Vec{z}_{t}^{k}=1}^{|\mathcal{Z}|}\prod_{i\neq n}q\left(z_{i,0:t}^{k}\right)\left[\ln q\left(z_{n,0:t}^{k}\right)\right] \\
&\rightarrow \sum_{k=1}^{K}\frac{1}{K}\sum_{t=0}^{T_k}\sum_{\Vec{z}_{0}^{k}...\Vec{z}_{t}^{k}=1}^{|\mathcal{Z}|}\prod_{i\neq n}q\left(z_{i,0:t}^{k}\right)\left[\int q(\Theta)\ln\Tilde{r}_{t}^{k}\prod_{n=1}^{N}p
\left(a_{n,0:t}^{k},z_{n,0:t}^{k}\vert o_{n,1:t}^{k},\Theta\right)d\Theta\right] \\
&=\sum_{k=1}^{K}\frac{1}{K}\sum_{t=0}^{T_k}\sum_{\Vec{z}_{0}^{k}...\Vec{z}_{t}^{k}=1}^{|\mathcal{Z}|}\prod_{i\neq n}q\left(z_{i,0:t}^{k}\right)\left[\ln q
\left(z_{n,0:t}^{k}\right)\right] \\
&\rightarrow \ln q\left(z_{n,0:t}^{k}\right) \\
&=\int q(\Theta)\ln\Tilde{r}_{t}^{k}\prod_{n=1}^{N}p\left(a_{n,0:t}^{k},z_{n,0:t}^{k}\vert o_{n,1:t}^{k},\Theta\right)d\Theta d\alpha \\
&=\text{E}_{q(\Theta)}\left[\ln\Tilde{r}_{t}^{k}\prod_{n=1}^{N}p\left(a_{n,0:t}^{k},z_{n,0:t}^{k}\vert o_{n,1:t}^{k},\Theta\right)\right]
\end{aligned}
\end{equation}
The optimal $q^*\left(z_{n,0:t}^{k}\right)$ has the form
\begin{equation}
\begin{aligned}
& q^*\left(z_{n,0:t}^{k}\right) \\
&\propto \exp\left\{\text{E}_{q(\Theta)}\left[\ln\Tilde{r}_{t}^{k}\prod_{n=1}^{N}p\left(a_{n,0:t}^{k},z_{n,0:t}^{k}\vert o_{n,1:t}^{k},\Theta\right)\right]\right\} \\
&=\exp\left\{\text{E}_{q(\Theta)}\left[\ln\Tilde{r}_{t}^{k}\right]+\sum_{n=1}^{N}\text{E}_{q(\Theta)}\left[\ln p\left(a_{n,0:t}^{k},z_{n,0:t}^{k}\vert o_{n,1:t}^{k},\Theta\right)\right]\right\}
\end{aligned}
\label{equ:q(z)}
\end{equation}
Due to the independence between agents, remove all term with indices other than $(n,k,t)$, the above equation is proportional to
\begin{equation}
\medmuskip=1mu
\thinmuskip=1mu
\thickmuskip=1mu
\begin{aligned}
&\exp\left\{\text{E}_{q(\Theta)}\left[\ln\Tilde{r}_{t}^{k}\right]+\text{E}_{q(\Theta)}\left[\ln p\left(a_{n,0:t}^{k},z_{n,0:t}^{k}\vert o_{n,1:t}^{k},\Theta\right)\right]\right\} \\
&=\exp\left\{\ln\Tilde{r}_{t}^{k}+\text{E}_{q(\Theta)}\left[\ln p\left(a_{n,0:t}^{k},z_{n,0:t}^{k}\vert o_{n,1:t}^{k},\Theta\right)\right]\right\} \\
&=\Tilde{r}_{t}^{k}\exp\left\{\text{E}_{q(\Theta)}\left[\ln \eta_n^{z_0}\pi_{n,z_0}^{k,a_0}\prod_{\tau=1}^{t}\omega_{n,a_{\tau-1},o_{\tau}}^{k,z_{\tau-1},z_{\tau}}\pi_{n,z_{\tau}}^{k,a_{\tau}}\right]\right\} \\
&=\Tilde{r}_{t}^{k}\exp\left\{\text{E}_{q(\Theta)}\left[\ln \eta_n^{z_0}+\sum_{\tau=0}^{t}\ln\pi_{n,z_{\tau}}^{k,a_{\tau}}+\sum_{\tau=1}^{t}\ln\omega_{n,a_{\tau-1},o_{\tau}}^{k,z_{\tau-1},z_{\tau}}\right]\right\} \\
&=\Tilde{r}_{t}^{k}\exp\left\{\text{E}_{q(u)}\left[\ln\eta_n^{z_0}\right]+\sum_{\tau=0}^{t}\text{E}_{q(\pi)}\left[\ln\pi_{n,z_{\tau}}^{k,a_{\tau}}\right]+\sum_{\tau=1}^{t}\text{E}_{q(V)}\left[\ln\omega_{n,a_{\tau-1},o_{\tau}}^{k,z_{\tau-1},z_{\tau}}\right]\right\} \\
&=\Tilde{r}_{t}^{k}\exp\left\{\text{E}_{q(u)}\left[\ln\eta_n^{z_0}\right]\right\}\prod_{\tau=0}^{t}\exp\left\{\text{E}_{q(\pi)}\left[\ln\pi_{n,z_{\tau}}^{k,a_{\tau}}\right]\right\}\prod_{\tau=1}^{t}\exp\left\{\text{E}_{q(V)}\left[\ln\omega_{n,a_{\tau-1},o_{\tau}}^{k,z_{\tau-1},z_{\tau}}\right]\right\} \\
&=\Tilde{r}_{t}^{k}\Tilde{\eta}_n^{z_0}\prod_{\tau=0}^{t}\Tilde{\pi}_{n,z_{n,\tau}^{k}}^{a_{n,\tau}^{k}}\prod_{\tau=1}^{t}\Tilde{\omega}_{n,a_{\tau-1},o_{\tau}}^{k,z_{\tau-1},z_{\tau}}
\end{aligned}
\label{equ:q(z)_result}
\end{equation}
Where $\Tilde{\Theta}_n=(\Tilde{\eta}_n, \Tilde{\omega}_{n}, \Tilde{\pi}_{n})$ and
\begin{equation}
\begin{aligned}
&\Tilde{\eta}_n^{z_0}=\exp\left\{\text{E}_{q(u)}\left[\ln\eta_n^{z_0}\right]\right\} \\
&\Tilde{\pi}_{n,z_{n,\tau}^{k}}^{a_{n,\tau}^{k}}=\exp\left\{\text{E}_{q(\pi)}\left[\ln\pi_{n,z_{\tau}}^{k,a_{\tau}}\right]\right\} =\exp\left\{\Psi\left(\phi_{n,z_{n,\tau}^{k}}^{a_{n,\tau}^{k}}\right)-\Psi\left(\sum_{a=1}^{|\mathcal{A}_n|}\phi_{n,z_{n,\tau}^{k}}^{a}\right)\right\} \\
&\Tilde{\omega}_{n,a_{\tau-1},o_{\tau}}^{k,z_{\tau-1},z_{\tau}}=\exp\left\{\text{E}_{q(V)}\left[\ln\omega_{n,a_{\tau-1},o_{\tau}}^{k,z_{\tau-1},z_{\tau}}\right]\right\}
\end{aligned}
\end{equation}
$\eta$ and $\omega$ are constructed by the stick-breaking process. For different destination node, the terms in exponential $\exp\{\cdot\}$ are computed by 
\begin{align}
&\text{E}_{q(u)}\left[\ln\eta_n^1\right]=\text{E}_{q(u)}\left[\ln u_n^1\right]=\Psi(\delta_n^1)-\Psi(\delta_n^1+\mu_n^1) \\
&\begin{aligned}
&\text{E}_{q(u)}\left[\ln\eta_n^i\right] \\
&=\text{E}_{q(u)}\left[\ln u_n^i\prod_{m=1}^{i-1}(1-u_n^m)\right] \\
&=\text{E}_{q(u)}\left[\ln u_n^i\right]+\sum_{m=1}^{i-1}\text{E}_{q(u)}\left[\ln (1-u_n^m)\right] \\
&=\Psi(\delta_n^i)-\Psi(\delta_n^i+\mu_n^i)+\sum_{m=1}^{i-1}\left[\Psi(\mu_n^m)-\Psi(\delta_n^m+\mu_n^m)\right] \ \text{for }i=2,...,|\mathcal{Z}_n|-1
\end{aligned} \\
&\text{E}_{q(u)}\left[\ln\eta_n^{|\mathcal{Z}_n|}\right]=\text{E}_{q(u)}\left[\ln \prod_{m=1}^{|\mathcal{Z}_n|-1}(1-u_n^m)\right]=\sum_{m=1}^{|\mathcal{Z}_n|-1}\left[\Psi(\mu_n^m)-\Psi(\delta_n^m+\mu_n^m)\right]
\end{align}
and
\begin{align}
&\begin{aligned}
&\text{E}_{q(V)}\left[\ln\omega_{n,a_{\tau-1},o_{\tau}}^{k,z_{\tau-1},1}\right] \\
&=\text{E}_{q(V)}\left[\ln V_{n,a_{\tau-1},o_{\tau}}^{k,z_{\tau-1},1}\right] \\ &=\Psi\left(\sigma_{n,a_{\tau-1},o_{\tau}}^{k,z_{\tau-1},1}\right)-\Psi\left(\sigma_{n,a_{\tau-1},o_{\tau}}^{k,z_{\tau-1},1}+\lambda_{n,a_{\tau-1},o_{\tau}}^{k,z_{\tau-1},1}\right)
\end{aligned} \\
&\begin{aligned}
&\text{E}_{q(V)}\left[\ln\omega_{n,a_{\tau-1},o_{\tau}}^{k,z_{\tau-1},i}\right] \\
&=\text{E}_{q(V)}\left[\ln V_{n,a_{\tau-1},o_{\tau}}^{k,z_{\tau-1},i}\prod_{m=1}^{i-1}\left(1-V_{n,a_{\tau-1},o_{\tau}}^{k,z_{\tau-1},m}\right)\right] \\
&=\Psi\left(\sigma_{n,a_{\tau-1},o_{\tau}}^{k,z_{\tau-1},i}\right)-\Psi\left(\sigma_{n,a_{\tau-1},o_{\tau}}^{k,z_{\tau-1},i}+\lambda_{n,a_{\tau-1},o_{\tau}}^{k,z_{\tau-1},i}\right) \\
&+\sum_{m=1}^{i-1}\left[\Psi\left(\lambda_{n,a_{\tau-1},o_{\tau}}^{k,z_{\tau-1},m}\right)-\Psi\left(\sigma_{n,a_{\tau-1},o_{\tau}}^{k,z_{\tau-1},m}+\lambda_{n,a_{\tau-1},o_{\tau}}^{k,z_{\tau-1},m}\right)\right] \ \text{for }i=2,...,|\mathcal{Z}_n|-1
\end{aligned} \\
&\begin{aligned}
&\text{E}_{q(V)}\left[\ln\omega_{n,a_{\tau-1},o_{\tau}}^{k,z_{\tau-1},{|\mathcal{Z}_n|}}\right] \\
&=\text{E}_{q(V)}\left[\ln \prod_{m=1}^{|\mathcal{Z}_n|-1}\left(1-V_{n,a_{\tau-1},o_{\tau}}^{k,z_{\tau-1},m}\right)\right] \\
&=\sum_{m=1}^{|\mathcal{Z}_n|-1}\left[\Psi\left(\lambda_{n,a_{\tau-1},o_{\tau}}^{k,z_{\tau-1},m}\right)-\Psi\left(\sigma_{n,a_{\tau-1},o_{\tau}}^{k,z_{\tau-1},m}+\lambda_{n,a_{\tau-1},o_{\tau}}^{k,z_{\tau-1},m}\right)\right]
\end{aligned}
\end{align}
In \ref{equ:q(z)}, the proportional expression is utilized to represent the $q\left(z_{n,0:t}^{k}\right)$. To make $q\left(z_{n,0:t}^{k}\right)$ proper distribution of $z_{n,0:t}^{k}$, i.e., satisfy \ref{equ:q(z)_constraint}, we re-write the final result in \ref{equ:q(z)_result} and substitute it into the constraint equation
\begin{equation}
\medmuskip=1mu
\thinmuskip=1mu
\thickmuskip=1mu
\begin{aligned}
&\frac{1}{K}\sum_{k,t}\sum_{z_{1:N,0}^k=1}^{|\mathcal{Z}|}\cdots\sum_{z_{1:N,t}^k=1}^{|\mathcal{Z}|}\prod_{n=1}^{N}q\left(z_{n,0:t}^{k}\right) \\
&=\frac{1}{K}\sum_{k,t}\sum_{z_{1:N,0}^k=1}^{|\mathcal{Z}|}\cdots\sum_{z_{1:N,t}^k=1}^{|\mathcal{Z}|}\prod_{n=1}^{N}\Tilde{r}_{t}^{k}\Tilde{\eta}_n^{z_0}\prod_{\tau=0}^{t}\Tilde{\pi}_{n,z_{n,\tau}^{k}}^{a_{n,\tau}^{k}}\prod_{\tau=1}^{t}\Tilde{\omega}_{n,a_{\tau-1},o_{\tau}}^{k,z_{\tau-1},z_{\tau}} \\
&=\frac{1}{K}\sum_{k,t}\Tilde{r}_{t}^{k}\sum_{z_{1:N,0}^k=1}^{|\mathcal{Z}|}\cdots\sum_{z_{1:N,t}^k=1}^{|\mathcal{Z}|}\prod_{n=1}^{N}p\left(a_{n,0:t}^k,z_{n,0:t}^k\vert o_{n,1:t}^k,\Tilde{\Theta}_n\right) \\
&=\frac{1}{K}\sum_{k,t}\Tilde{r}_{t}^{k}\sum_{z_{1:N,0}^k=1}^{|\mathcal{Z}|}\cdots\sum_{z_{1:N,t}^k=1}^{|\mathcal{Z}|}\prod_{n=1}^{N}p\left(a_{n,0:t}^k\vert o_{n,1:t}^k,\Tilde{\Theta}_n\right)p\left(z_{n,0:t}^k\vert a_{n,0:t}^k,o_{n,1:t}^k,\Tilde{\Theta}_n\right) \\
&=\frac{1}{K}\sum_{k,t}\Tilde{r}_{t}^{k}\sum_{z_{1:N,0}^k=1}^{|\mathcal{Z}|}\cdots\sum_{z_{1:N,t}^k=1}^{|\mathcal{Z}|}\prod_{n=1}^{N}p\left(a_{n,0:t}^k\vert o_{n,1:t}^k,\Tilde{\Theta}_n\right)\prod_{n=1}^{N}p\left(z_{n,0:t}^k\vert a_{n,0:t}^k,o_{n,1:t}^k,\Tilde{\Theta}_n\right) \\
&=\frac{1}{K}\sum_{k,t}\Tilde{r}_{t}^{k}\prod_{n=1}^{N}p\left(a_{n,0:t}^k\vert o_{n,1:t}^k,\Tilde{\Theta}_n\right)\sum_{z_{1:N,0}^k=1}^{|\mathcal{Z}|}\cdots\sum_{z_{1:N,t}^k=1}^{|\mathcal{Z}|}\prod_{n=1}^{N}p\left(z_{n,0:t}^k\vert a_{n,0:t}^k,o_{n,1:t}^k,\Tilde{\Theta}_n\right) \\
&=\frac{1}{K}\sum_{k,t}\Tilde{r}_{t}^{k}\prod_{n=1}^{N}p\left(a_{n,0:t}^k\vert o_{n,1:t}^k,\Tilde{\Theta}_n\right)
\end{aligned}
\end{equation}
By \ref{def: emp_likelihood} and $\Tilde{r}_t^k=\gamma^t\frac{r_t^k-R_{\text{min}}}{\prod_{n=1}^{N}p(a_{n,0:t}^k|o_{n,1:t}^k,\Pi)}$, the above result is just equal to $\hat{V}(D^K;\Tilde{\Theta})$. Thus,
\begin{equation}
\medmuskip=1mu
\thinmuskip=1mu
\thickmuskip=1mu
\begin{aligned}
&\frac{1}{K}\sum_{k,t}\frac{\Tilde{r}_{t}^{k}\prod_{n=1}^{N}p\left(a_{n,0:t}^k\vert o_{n,1:t}^k,\Tilde{\Theta}_n\right)}{\hat{V}(D^K;\Tilde{\Theta})}\sum_{z_{1:N,0}^k=1}^{|\mathcal{Z}|}\cdots\sum_{z_{1:N,t}^k=1}^{|\mathcal{Z}|}\prod_{n=1}^{N}p\left(z_{n,0:t}^k\vert a_{n,0:t}^k,o_{n,1:t}^k,\Tilde{\Theta}_n\right) \\
&=\frac{1}{K}\sum_{k,t}\Tilde{\nu}_{t}^{k}\sum_{z_{1:N,0}^k=1}^{|\mathcal{Z}|}\cdots\sum_{z_{1:N,t}^k=1}^{|\mathcal{Z}|}\prod_{n=1}^{N}p\left(z_{n,0:t}^k\vert a_{n,0:t}^k,o_{n,1:t}^k,\Tilde{\Theta}_n\right) \\
&=\frac{1}{K}\sum_{k,t}\sum_{z_{1:N,0}^k=1}^{|\mathcal{Z}|}\cdots\sum_{z_{1:N,t}^k=1}^{|\mathcal{Z}|}\prod_{n=1}^{N}\Tilde{\nu}_{t}^{k}p\left(z_{n,0:t}^k\vert a_{n,0:t}^k,o_{n,1:t}^k,\Tilde{\Theta}_n\right) \\
&=\frac{1}{K}\sum_{k,t}\sum_{z_{1:N,0}^k=1}^{|\mathcal{Z}|}\cdots\sum_{z_{1:N,t}^k=1}^{|\mathcal{Z}|}\prod_{n=1}^{N}q\left(z_{n,0:t}^k\right) \\
&=1
\end{aligned}
\end{equation}
$\Tilde{\nu}_{t}^{k}$ is the reweighted reward that makes $q\left(z_{n,0:t}^k\right)$ satisfy \ref{equ:q(z)_constraint}.

For optimal $q^*(\Theta_n)$, use formula in \ref{equ: VIoptformula}, keep all other $q$ distributions fixed and treat terms unrelated to $q(\Theta_n)$ as constants, the result can be obtained as
\begin{equation}
\medmuskip=0mu
\thinmuskip=0mu
\thickmuskip=0mu
\begin{aligned}
&q^*(\Theta_n) \\
&\propto \exp\left\{\text{E}_{q(z,\rho,\alpha)}\left[\ln\Tilde{r}_{t}^{k}p(a_{n,0:t}^{k},z_{n,0:t}^{k}\vert o_{n,1:t}^{k},\Theta)p(\Theta_n)p(\rho_n)p(\alpha_n)\right]\right\} \\
&\propto \exp\left\{\text{E}_{q(z,\rho,\alpha)}\left[\ln\Tilde{r}_{t}^{k}p(a_{n,0:t}^{k},z_{n,0:t}^{k}\vert o_{n,1:t}^{k},\Theta)p(\Theta_n)\right]\right\} \\
&=\exp\left\{\text{E}_{q(z)}\left[\ln\Tilde{r}_{t}^{k}p(a_{n,0:t}^{k},z_{n,0:t}^{k}\vert o_{n,1:t}^{k},\Theta)\right]+\text{E}_{q(\rho,\alpha)}\left[\ln p(\Theta_n)\right]\right\} \\
&=\exp\left\{\text{E}_{q(z)}\left[\ln\Tilde{r}_{t}^{k}\eta_n^{z_0}\pi_{n,z_0}^{k,a_0}\prod_{\tau=1}^{t}\omega_{n,a_{\tau-1},o_{\tau}}^{k,z_{\tau-1},z_{\tau}}\pi_{n,z_{\tau}}^{k,a_{\tau}}\right]+\text{E}_{q(\rho,\alpha)}\left[\ln p(u_n\vert\rho_n)p(V_n\vert\alpha_n)p(\pi_n)\right]\right\} \\
&\propto \exp\Biggl\{\frac{1}{K}\sum_{k,t,z_{n,0:t}^{k}}q(z_{n,0:t}^{k})\left[\ln\eta_n^{z_0}+\sum_{\tau=0}^{t}\ln\pi_{n,z_{\tau}}^{k,a_{\tau}}+\sum_{\tau=1}^{t}\ln\omega_{n,a_{\tau-1},o_{\tau}}^{k,z_{\tau-1},z_{\tau}}\right] \\
&+\text{E}_{q(\rho)}\left[\ln p(u_n\vert\rho_n)\right]+\text{E}_{q(\alpha)}\left[\ln p(V_n\vert\alpha_n)\right]+\ln p(\pi_n)\Biggr\} \\
&=\exp\left\{\left[\frac{1}{K}\sum_{k,t,z_{n,0:t}^{k}}q(z_{n,0:t}^{k})\left[\ln\eta_n^{z_0}\right]+\text{E}_{q(\rho)}\left[\ln p(u_n\vert\rho_n)\right]\right]\right. \\
&+\left[\frac{1}{K}\sum_{k,t,z_{n,0:t}^{k}}q(z_{n,0:t}^{k})\left[\sum_{\tau=1}^{t}\ln\omega_{n,a_{\tau-1},o_{\tau}}^{k,z_{\tau-1},z_{\tau}}\right]+\text{E}_{q(\alpha)}\left[\ln p(V_n\vert\alpha_n)\right]\right] \\
&+\left.\left[\frac{1}{K}\sum_{k,t,z_{n,0:t}^{k}}q(z_{n,0:t}^{k})\left[\sum_{\tau=0}^{t}\ln\pi_{n,z_{\tau}}^{k,a_{\tau}}\right]+\ln p(\pi_n)\right]\right\} \\
&=\exp\left\{\left[\frac{1}{K}\sum_{k,t,z_{n,0:t}^{k}}q(z_{n,0:t}^{k})\left[\ln u_n^{z_0}\prod_{m=1}^{z_0-1}(1-u_n^m)\right]+\text{E}_{q(\rho)}\left[\ln p(u_n\vert\rho_n)\right]\right]\right. \\
&+\left[\frac{1}{K}\sum_{k,t,z_{n,0:t}^{k}}q(z_{n,0:t}^{k})\sum_{\tau=1}^{t}\left[\ln V_{n,a_{\tau-1},o_{\tau}}^{k,z_{\tau-1},z_{\tau}}\prod_{m=1}^{z_{\tau}-1}\left(1-V_{n,a_{\tau-1},o_{\tau}}^{k,z_{\tau-1},m}\right)\right]+\text{E}_{q(\alpha)}\left[\ln p(V_n\vert\alpha_n)\right]\right] \\
&+\left.\left[\frac{1}{K}\sum_{k,t,z_{n,0:t}^{k}}q(z_{n,0:t}^{k})\left[\sum_{\tau=0}^{t}\ln\pi_{n,z_{\tau}}^{k,a_{\tau}}\right]+\ln p(\pi_n)\right]\right\}
\end{aligned}
\end{equation}
In above formula, there are three parts of variables in the exponential term,
{
\medmuskip=0mu
\thinmuskip=0mu
\thickmuskip=0mu
\begin{align}
&\begin{aligned}
&\frac{1}{K}\sum_{k,t,z_{n,0:t}^{k}}q(z_{n,0:t}^{k})\left[\ln u_n^{z_0}\prod_{m=1}^{z_0-1}(1-u_n^m)\right]+\text{E}_{q(\rho)}\left[\ln p(u_n\vert\rho_n)\right] \\
&=\frac{1}{K}\sum_{k,t,z_{n,0:t}^{k}}q(z_{n,0:t}^{k})\left[\ln u_n^{z_0}+\sum_{m=1}^{z_0-1}\ln(1-u_n^m)\right]+\text{E}_{q(\rho)}\left[\ln p(u_n\vert\rho_n)\right]
\end{aligned} \label{equ:eta_update} \\
&\begin{aligned}
&\frac{1}{K}\sum_{k,t,z_{n,0:t}^{k}}q(z_{n,0:t}^{k})\sum_{\tau=1}^{t}\left[\ln V_{n,a_{\tau-1},o_{\tau}}^{k,z_{\tau-1},z_{\tau}}\prod_{m=1}^{z_{\tau}-1}V_{n,a_{\tau-1},o_{\tau}}^{k,z_{\tau-1},m}\right]+\text{E}_{q(\alpha)}\left[\ln p(V_n\vert\alpha_n)\right] \\
&=\frac{1}{K}\sum_{k,t,z_{n,0:t}^{k}}q(z_{n,0:t}^{k})\sum_{\tau=1}^{t}\left[\ln V_{n,a_{\tau-1},o_{\tau}}^{k,z_{\tau-1},z_{\tau}}+\sum_{m=1}^{z_{\tau}-1}\ln\left(1- V_{n,a_{\tau-1},o_{\tau}}^{k,z_{\tau-1},m}\right)\right]+\text{E}_{q(\alpha)}\left[\ln p(V_n\vert\alpha_n)\right]
\end{aligned} \label{equ:omega_update} \\
& \frac{1}{K}\sum_{k,t,z_{n,0:t}^{k}}q(z_{n,0:t}^{k})\left[\sum_{\tau=0}^{t}\ln\pi_{n,z_{\tau}}^{k,a_{\tau}}\right]+\ln p(\pi_n) \label{equ:pi_update}
\end{align}
}
By the conjugacy between prior and likelihood models, we know each $q$ distribution belongs to the same family of its corresponding prior and they are all in exponential family, thus the computation of $q$ distribution can reduce to the computation of its parameters in their exponential expression. By re-positioning components in above equations in terms of each variable, the parameters for each $q$ distribution can be computed.

For $u_n^i$, re-write the prior and variational distributions in terms of exponential family,
\begin{equation}
\begin{aligned}
& \text{E}_{q_(\rho)}\left[\ln p(u_n^i\vert\rho_n)\right]\propto (1-1)\ln u_n^i+\left(\text{E}_{q_(\rho)}\left[\rho_n\right]-1\right)\ln(1-u_n^i) \\
& \ln q(u_n^i)\propto (\delta_n^i-1)\ln u_n^i+(\mu_n^i-1)\ln(1-u_n^i)
\end{aligned}
\end{equation}
For $\ln u_n^i$, only $(z_{n,0}^k=i)$ is associated with it and all cases with indices $m>i$ must be collected for $\ln(1-\ln u_n^i)$; we rearrange components in \ref{equ:eta_update} and obtain
\begin{align}
&\begin{aligned}
&\left[\frac{1}{K}\sum_{k,t,z_{n,0:t}^{k}}q_{n,t}^k(z_{n,0}^{k}=i)\right]\ln u_n^i=(\delta_n^i-1)\ln u_n^i \\
&\rightarrow \delta_n^i=1+\frac{1}{K}\sum_{k,t,z_{n,0:t}^{k}}q_{n,t}^k(z_{n,0}^{k}=i)
\end{aligned} \\
&{
\medmuskip=1mu
\thinmuskip=1mu
\thickmuskip=1mu
\begin{aligned}
&\left[\sum_{m=i+1}^{|\mathcal{Z}_n|-1}\frac{1}{K}\sum_{k,t,z_{n,0:t}^{k}}q_{n,t}^k(z_{n,0}^{k}=m)+\text{E}_{q_(\rho)}\left[\rho_n\right]-1\right]\ln(1-u_n^i)=(\mu_n^i-1)\ln(1-u_n^i) \\
&\rightarrow \mu_n^i=\frac{g_n}{h_n}+\sum_{m=i+1}^{|\mathcal{Z}_n|}\frac{1}{K}\sum_{k,t,z_{n,0:t}^{k}}q_{n,t}^k(z_{n,0}^{k}=i)
\end{aligned}
}
\end{align}
Where $\text{E}_{q_(\rho)}\left[\rho_n\right]=\frac{g_n}{h_n}$.

The update of $q\left(V_{n,a,o}^{i,j}\right)$ is similar to the update of $q(u_n^i)$. Rewrite $q(V_{n,a,o}^{i,j})$ and $\text{E}_{q(\alpha)}\left[p\left(V_{n,a,o}^{i,j}\vert\alpha_{n,a,o}^i\right)\right]$ in terms of exponential family,
\begin{equation}
\medmuskip=1mu
\thinmuskip=1mu
\thickmuskip=1mu
\begin{aligned}
&\text{E}_{q(\alpha)}\left[\ln p\left(V_{n,a,o}^{i,j}\vert\alpha_{n,a,o}^{i}\right)\right]\propto(1-1)\ln V_{n,a,o}^{i,j}+\left(\text{E}_{q(\alpha)}\left[\alpha_{n,a,o}^{i}\right]-1\right)\ln\left(1-V_{n,a,o}^{i,j}\right) \\
&\ln q\left(V_{n,a,o}^{i,j}\right)\propto \left(\sigma_{n,a,o}^{i,j}-1\right)\ln V_{n,a,o}^{i,j}+\left(\lambda_{n,a,o}^{i}-1\right)\ln\left(1-V_{n,a,o}^{i,j}\right)
\end{aligned}
\end{equation}
Since only case $\left(z_{n,\tau-1}^{k}=i,z_{n,\tau}^{k}=j\right)$ associated with $\ln V_{n,a,o}^{i,j}$, rearrange terms related to it,
\begin{equation}
\begin{aligned}
&\left[\sum_{k,t}\frac{1}{K}\sum_{\tau=1}^{t}q_{n,t}^k(z_{n,\tau-1}^{k}=i,z_{n,\tau}^{k}=j)\mathbb{I}(a_{n,\tau-1}^{k}=a,o_{n,\tau}^{k}=o)\right]\ln V_{n,a,o}^{i,j} \\
&=\left(\sigma_{n,a,o}^{i,j}-1\right)\ln V_{n,a,o}^{i,j} \\
&\rightarrow \sigma_{n,a,o}^{i,j}=1+\sum_{k,t}\frac{1}{K}\sum_{\tau=1}^{t}q_{n,t}^k(z_{n,\tau-1}^{k}=i,z_{n,\tau}^{k}=j)\mathbb{I}(a_{n,\tau-1}^{k}=a,o_{n,\tau}^{k}=o)
\end{aligned}
\end{equation}
For $\ln\left(1-V_{n,a,o}^{i,j}\right)$, all cases $\left(z_{n,\tau-1}^{k}=i,z_{n,\tau}^{k}=m\right)$ for $m>j$ must be considered, 
\begin{equation}
\medmuskip=0mu
\thinmuskip=0mu
\thickmuskip=0mu
\begin{aligned}
&\left[\sum_{m=j+1}^{|\mathcal{Z}_n|}\frac{1}{K}\sum_{k,t}\sum_{\tau=1}^{t}q_{n,t}^k(z_{n,\tau-1}^{k}=i,z_{n,\tau}^{k}=m)\mathbb{I}(a_{n,\tau-1}^{k}=a,o_{n,\tau}^{k}=o)+\text{E}_{q(\alpha)}\left[\alpha_{n,a,o}^{i}\right]-1\right]\ln\left(1-V_{n,a,o}^{i,j}\right) \\
&=\left(\lambda_{n,a,o}^{i,j}-1\right)\ln\left(1-V_{n,a,o}^{i,j}\right) \\
&\rightarrow \lambda_{n,a,o}^{i,j}=\frac{a_{n,a,o}^{i}}{b_{n,a,o}^{i}}+\sum_{m=j+1}^{|\mathcal{Z}_n|}\frac{1}{K}\sum_{k,t}\sum_{\tau=1}^{t}q_{n,t}^k(z_{n,\tau-1}^{k}=i,z_{n,\tau}^{k}=m)\mathbb{I}(a_{n,\tau-1}^{k}=a,o_{n,\tau}^{k}=o)
\end{aligned}
\end{equation}
Where $\text{E}_{q(\alpha)}\left[\alpha_{n,a,o}^{i}\right]=\frac{a_{n,a,o}^{i}}{b_{n,a,o}^{i}}$.

For the update of each $q(\pi_{n,i})$, rearrange components in \ref{equ:pi_update} in terms of the $q(\pi_{n,i})$ distribution,
\begin{equation}
\begin{aligned}
&\frac{1}{K}\sum_{k,t,z_{n,0:t}^{k}}q(z_{n,0:t}^{k})\sum_{\tau=0}^{t}\ln\pi_{n,z_{\tau}}^{k,a_{\tau}}+\ln p(\pi_{n}) \\
&=\frac{1}{K}\sum_{k,t,z_{n,0:t}^{k}}q(z_{n,0:t}^{k})\sum_{\tau=0}^{t}\ln\pi_{n,z_{\tau}}^{k,a_{\tau}}+\ln \prod_{i=1}^{|\mathcal{Z}_n|}\prod_{a=1}^{|\mathcal{A}_n|}\left(\pi_{n,i}^{a}\right)^{\theta_{n,i}^{a}-1} \\
&=\frac{1}{K}\sum_{k,t,z_{n,0:t}^{k}}q(z_{n,0:t}^{k})\sum_{\tau=0}^{t}\ln\pi_{n,z_{\tau}}^{k,a_{\tau}}+ \sum_{i=1}^{|\mathcal{Z}_n|}\sum_{a=1}^{|\mathcal{A}_n|}\left(\theta_{n,i}^{a}-1\right)\ln\pi_{n,i}^{a} \\
&=\sum_{i=1}^{|\mathcal{Z}_n|}\sum_{a=1}^{|\mathcal{A}_n|}\left[\theta_{n,i}^{a}+\frac{1}{K}\sum_{k,t}\sum_{\tau=0}^{t}q_{n,t}^k(z_{n,\tau}^{k}=i)\mathbb{I}(a_{n,\tau}^{k}=a)-1\right]\ln\pi_{n,i}^{a} \\
&=\sum_{i=1}^{|\mathcal{Z}_n|}\sum_{a=1}^{|\mathcal{A}_n|}\left(\phi_{n,i}^{a}-1\right)\ln\pi_{n,i}^{a} \\
&=\ln\prod_{i=1}^{|\mathcal{Z}_n|}\prod_{a=1}^{|\mathcal{A}_n|}\left(\pi_{n,i}^{a}\right)^{\phi_{n,i}^{a}-1} \\
&=\ln q(\pi_{n}) \\
&\rightarrow \phi_{n,i}^{a}=\theta_{n,i}^{a}+\frac{1}{K}\sum_{k,t}\sum_{\tau=0}^{t}q_{n,t}^k(z_{n,\tau}^{k}=i)\mathbb{I}(a_{n,\tau}^{k}=a)
\end{aligned}
\end{equation}
In the optimal formulas for $q(\Theta_n)$, we need $q_{n,t}^k(z_{n,\tau}^{k})=\Tilde{\nu}_t^k p(z_{n,\tau}^{k}|a_{n,0:t}^{k},o_{n,1:t}^{k},\Tilde{\Theta})$, where $p(z_{n,\tau}^{k}|a_{n,0:t}^{k},o_{n,1:t}^{k},\Tilde{\Theta})$ is the marginal distribution of $p(z_{n,0:t}^{k}|a_{n,0:t}^{k},o_{n,1:t}^{k},\Tilde{\Theta})$. Obtaining it by directly marginalizing the following joint distribution is extremely computationally cumbersome, 
\begin{equation}
\begin{aligned}
    & p(z_{n,\tau}^{k}|a_{n,0:t}^{k},o_{n,1:t}^{k},\Tilde{\Theta}) \\
    & =\sum_{z_{n,\forall t\neq\tau}^{k}=1}^{|\mathcal{Z}_n|}p(z_{n,0:t}^{k}|a_{n,0:t}^{k},o_{n,1:t}^{k},\Tilde{\Theta}) \\
    & =\sum_{z_{n,\forall t\neq\tau}^{k}=1}^{|\mathcal{Z}_n|}\frac{p(a_{n,0:t}^{k},z_{n,0:t}^{k}|o_{n,1:t}^{k},\Tilde{\Theta})}{p(a_{n,0:t}^{k}|o_{n,1:t}^{k},\Tilde{\Theta})} \\
    & p(a_{n,0:t}^{k}|o_{n,1:t}^{k},\Tilde{\Theta})=\sum_{z_{n,0:t}^{k}=1}^{|\mathcal{Z}_n|}p(a_{n,0:t}^{k},z_{n,0:t}^{k}|o_{n,1:t}^{k},\Tilde{\Theta})
\end{aligned}
\end{equation}
Instead, each marginal distribution for $\tau=0,...,t$ can be computed analytically by iterative method. The marginal distribution for each $\tau$ can be factorized into two independent sections according to the d-separation property of Bayes network,
\begin{equation}
\begin{aligned}
    & p(z_{n,\tau}^{k}=i|a_{n,0:t}^{k},o_{n,1:t}^{k},\Tilde{\Theta}) \\
    & \propto p(a_{n,0:t}^{k},z_{n,\tau}^{k}=i|o_{n,1:t}^{k},\Tilde{\Theta}) \\
    & =p(a_{n,0:\tau}^{k},z_{n,\tau}^{k}=i|o_{n,1:\tau}^{k},\Tilde{\Theta})p(a_{n,\tau+1:t}^{k}|z_{n,\tau}^{k}=i,a_{n,\tau:t}^k,o_{n,\tau+1:t}^{k},\Tilde{\Theta}) \\
    & =\alpha_{n,\tau}^k(i)\beta_{n,\tau}^{k,t}(i),
\end{aligned}
\end{equation}
where $\alpha$ and $\beta$ are similar to the forward-backward messages in hidden Markov models. For notational simplicity, we remove $\Tilde{\Theta}$ in the derivation of $\alpha$ and $\beta$. The $\alpha$ and $\beta$ can be computed recursively via dynamic programming,
\begin{equation}
\begin{aligned}
& \alpha_{n,\tau}^k(i) = p(a_{n,0:\tau}^{k},z_{n,\tau}^{k}=i|o_{n,1:\tau}^{k},\Tilde{\Theta}) \\
    & =
    \begin{cases}
     \eta_{n}^i\pi(a_{n,0}^k|z_{n,0}^k=i) & \tau = 0 \\
     \sum_{j=1}^{|\mathcal{z}_n|}\alpha_{n,\tau-1}^k(j)\omega(z_{n,\tau}^k=i|z_{n,\tau-1}^k=j,a_{n,\tau-1}^k,o_{n,\tau}^k)\pi(a_{n,\tau}^k|z_{n,\tau}^k=i) & \tau > 0
    \end{cases}
\end{aligned}
\end{equation}
\begin{equation}
\medmuskip=1mu
\thinmuskip=1mu
\thickmuskip=1mu
\begin{aligned}
& \beta_{n,\tau}^{k,t}(i) = p(a_{n,\tau+1:t}^{k}|z_{n,\tau}^{k}=i,a_{n,\tau}^{k},o_{n,\tau+1:t}^{k},\Tilde{\Theta}) \\
    & =
    \begin{cases}
     1 & \tau = t \\
     \sum_{j=1}^{|\mathcal{z}_n|}\omega(z_{n,\tau+1}^k=j|z_{n,\tau}^k=i,a_{n,\tau}^k,o_{n,\tau+1}^k)\pi(a_{n,\tau+1}^k|z_{n,\tau+1}^k=j)\beta_{n,\tau+1}^{k,t}(j) & \tau < t
    \end{cases}
\end{aligned}
\end{equation}
So the marginal distributions in $q(\Theta)$ update are computed by
\begin{align}
& p(z_{n,\tau}^k=i|a_{n,0:t}^k,o_{n,1:t}^k,\Tilde{\Theta})=\frac{\alpha_{n,\tau}^k(i)\beta_{n,\tau}^{k,t}(i)}{\sum_{i=1}^{|\mathcal{z}_n|}\alpha_{n,\tau}^k(i)\beta_{n,\tau}^{k,t}(i)} \\
&\begin{aligned}
    & p(z_{n,\tau-1}^k=i,z_{n,\tau}^k=j|a_{n,0:t}^k,o_{n,1:t}^k,\Tilde{\Theta}) \\ &=\frac{\alpha_{n,\tau-1}^k(i)\omega(z_{n,\tau}^k=j|z_{n,\tau-1}^k=i,a_{n,\tau-1}^k,o_{n,\tau}^k)\pi(a_{n,\tau}^k|z_{n,\tau}^k=j)\beta_{n,\tau}^{k,t}(j)}{\sum_{i,j=1}^{|\mathcal{z}_n|}\alpha_{n,\tau-1}^k(i)\omega(z_{n,\tau}^k=j|z_{n,\tau-1}^k=i,a_{n,\tau-1}^k,o_{n,\tau}^k)\pi(a_{n,\tau}^k|z_{n,\tau}^k=j)\beta_{n,\tau}^{k,t}(j)}
\end{aligned}
\end{align}

For the update of $q(\rho_n)$, start with the optimal formula and treat all components unrelated to $\rho_n$ as constant,
\begin{equation}
\medmuskip=1mu
\thinmuskip=1mu
\thickmuskip=1mu
\begin{aligned}
&q(\rho_n) \\
&\propto \exp\Biggl\{\text{E}_{q(\Theta,\alpha,z)}\left[\frac{1}{K}\sum_{k,t,z_{n,0:t}^{k}}\ln\Tilde{r}_{t}^{k} p\left(a_{n,0:t}^{k},z_{n,0:t}^{k}\vert o_{n,1:t}^{k},\Tilde{\Theta}\right)\right] \\
&+\text{E}_{q(\Theta,\alpha,z)}\left[\ln p(\Theta\vert\alpha_n, \rho_n)\right]+\text{E}_{q(\Theta,\alpha,z)}\left[\ln p(\rho_n)\right]\Biggr\} \\
&\propto \exp\left\{\text{E}_{q(\Theta,\alpha,z)}\left[\ln p(\Theta\vert\alpha_n, \rho_n)\right]+\text{E}_{q(\Theta,\alpha,z)}\left[\ln p(\rho_n)\right]\right\} \\
&\propto \exp\left\{\text{E}_{q(u)}\left[\ln p(u_n\vert\rho_n)\right]+\ln p(\rho_n)\right\} \\
&=\exp\left\{\text{E}_{q(u)}\left[\ln \prod_{i=1}^{|\mathcal{Z}_n|}p(u_n^i\vert\rho_n)\right]\right\}p(\rho_n) \\
&=\exp\left\{\text{E}_{q(u)}\left[\sum_{i=1}^{|\mathcal{Z}_n|}\ln \frac{\Gamma(1+\rho_n)}{\Gamma(1)\Gamma(\rho_n)}{u_n^i}^{1-1}(1-u_n^i)^{\rho_n-1}\right]\right\}\frac{f^e}{\Gamma(e)}\rho_n^(e-1)\exp\left\{-f\rho_n\right\} \\
&\propto \exp\left\{\sum_{i=1}^{|\mathcal{Z}_n|}\ln\rho_n+(e-1)\ln\rho_n+(\rho_n-1)\sum_{i=1}^{|\mathcal{Z}_n|}\text{E}_{q(u)}\left[\ln(1-u_n^i)\right]-f\rho_n\right\} \\
&=\exp\left\{(e+|\mathcal{Z}_n|-1)\ln\rho_n+(\rho_n-1)\sum_{i=1}^{|\mathcal{Z}_n|}\left[\Psi\left(\mu_n^i\right)-\Psi\left(\delta_n^i+\mu_n^i\right)\right]-f\rho_n\right\} \\
&=\rho_n^{-1}\exp\left\{(e+|\mathcal{Z}_n|)\ln\rho_n-\rho_n\left(f-\sum_{i=1}^{|\mathcal{Z}_n|}\left[\Psi\left(\mu_n^i\right)-\Psi\left(\delta_n^i+\mu_n^i\right)\right]\right)+C_{\rho_n}\right\} \\
&\approx \text{Gamma}(g_n,h_n)
\end{aligned}
\end{equation}
Since $q(\rho_n)$ is assumed to be Gamma distribution, compare the above expression with $\text{Gamma}(g_n,h_n)$, we can obtain
\begin{equation}
\begin{aligned}
& g_n = e+|\mathcal{Z}_n| \\
& h_n = f-\sum_{i=1}^{|\mathcal{Z}_n|}\left[\Psi\left(\mu_n^i\right)-\Psi\left(\delta_n^i+\mu_n^i\right)\right]
\end{aligned}
\end{equation}

For the update of each $q(\alpha_{n,a,o}^{i})$, from optimal formula we have
\begin{equation}
\medmuskip=1mu
\thinmuskip=1mu
\thickmuskip=1mu
\begin{aligned}
&q(\alpha_{n}) \\
&\propto \exp\Biggl\{\text{E}_{q(\Theta,\rho,z)}\left[\frac{1}{K}\sum_{k,t,z_{n,0:t}^{k}}\ln\Tilde{r}_{t}^{k} p(a_{n,0:t}^{k},z_{n,0:t}^{k}\vert o_{n,1:t}^{k},\Theta)\right] \\
&+\text{E}_{q(\Theta,\rho,z)}\left[\ln p(\Theta_n\vert\alpha_n)\right]+\text{E}_{q(\Theta,\rho,z)}\left[\ln p(\alpha_n)\right]\Biggr\} \\
&\propto \exp\left\{\text{E}_{q(\Theta,\rho,z)}\left[\ln p(\Theta_n\vert\alpha_n)\right]+\text{E}_{q(\Theta,\rho,z)}\left[\ln p(\alpha_n)\right]\right\} \\
&\propto\exp\left\{\text{E}_{q(V)}\left[\ln p(V_n\vert\alpha_n)\right]+\ln p(\alpha_n)\right\} \\
&=\exp\left\{\text{E}_{q(V)}\left[\ln p(V_n\vert\alpha_n)\right]\right\}p(\alpha_n) \\
&\text{(for each $\alpha$ with }(n,a,o,i)\text{ indices)} \\
&\rightarrow \exp\left\{\text{E}_{q(V)}\left[\ln\prod_{j=1}^{|\mathcal{Z}_n|}p(V_{j}\vert\alpha)\right]\right\}p(\alpha) \\
&=\exp\left\{\sum_{j=1}^{|\mathcal{Z}_n|}\text{E}_{q(V)}\left[\ln p(V_{j}\vert\alpha)\right]\right\}p(\alpha) \\
&=\exp\left\{\sum_{j=1}^{|\mathcal{Z}_n|}\text{E}_{q(V)}\left[\ln\frac{\Gamma(1+\alpha)}{\Gamma(1)\Gamma(\alpha)}V_j^{1-1}(1-V_j)^{\alpha-1}\right]\right\}\frac{d^{c}}{\Gamma(c)}\alpha^{c-1}\exp\left\{-d\alpha\right\} \\
&\propto \exp\left\{\sum_{j=1}^{|\mathcal{Z}_n|}\ln\alpha+(\alpha-1)\sum_{j=1}^{|\mathcal{Z}_n|}\text{E}_{q(V)}\left[\ln (1-V_j)\right]+(c-1)\ln\alpha-d\alpha\right\} \\
&=\exp\left\{\left(c+|\mathcal{Z}_n|-1\right)\ln\alpha+(\alpha-1)\sum_{j=1}^{|\mathcal{Z}_n|}\left[\Psi(\lambda_j)-\Psi(\sigma_j+\lambda_j)\right]-d\alpha\right\} \\
&=\alpha^{-1}\exp\left\{\left(c+|\mathcal{Z}_n|\right)\ln\alpha+\alpha\left(d-\sum_{j=1}^{|\mathcal{Z}_n|}\left[\Psi(\lambda_j)-\Psi(\sigma_j+\lambda_j)\right]\right)-C_{\alpha}\right\} \\
&\approx \text{Gamma}(a,b)
\end{aligned}
\end{equation}
Similar to $q(\rho_n)$, $q(\alpha)$ is Gamma distribution with parameters $(a,b)$, apply the above optimal formula to all $q(\alpha_{n,a,o}^i)$, the derivation of each $q(\alpha_{n,a,o}^i)$ can be obtained by the following update,
\begin{equation}
\begin{aligned}
& a_{n,a,o}^{i}=c_{n,a,o}+|\mathcal{Z}_n| \\
& b_{n,a,o}^{i}=d_{n,a,o}-\sum_{j=1}^{|\mathcal{Z}_n|}\left[\Psi\left(\lambda_{n,a,o}^{i,j}\right)-\Psi\left(\sigma_{n,a,o}^{i,j}+\lambda_{n,a,o}^{i,j}\right)\right]
\end{aligned}
\end{equation}
\end{appendices}
\begin{appendices}
\chapter{DISTRIBUTIONS OF RANDOM VARIABLES}

In this work Beta, Gamma, and Dirichlet distributions are utilized for prior models; their equations are presented here.
\begin{defn}
A continuous random variable $V$ is Beta distributed with parameters $(\sigma, \lambda)$ if its probability density function $p(V)$ has the following form:
\[
\begin{aligned}
    & V\sim\textup{Beta}(\sigma,\lambda) \\ 
    & p(V)=\frac{\Gamma(\sigma+\lambda)}{\Gamma(\sigma)\Gamma(\lambda)}V^{\sigma-1}(1-V)^{\lambda-1}
\end{aligned}
\]
\end{defn}
$\Gamma(\cdot)$ is the Gamma function and $\Gamma(n)=(n-1)!$. The realization of $V$ is within range $[0,1]$, so sample of $V$ can be taken as a probability value. Beta distribution is the conjugate prior of Bernoulli and Binomial distributions.
\begin{defn}
A continuous random variable $\alpha$ possesses Gamma distribution with parameters $(c,d)$ if its probability density function $p(\alpha)$ is described as
\[
\begin{aligned}
    & \alpha\sim\textup{Gamma}(c,d)\\
    & p(\alpha)=\frac{d^c}{\Gamma(c)}\alpha^{c-1}e^{-\alpha d}    
\end{aligned}
\]
\end{defn}
The support of $\alpha$ is positive real numbers $(0,\infty)$. Gamma distribution is the conjugate prior of Poisson distribution.
Dirichlet distribution is generalization of Beta distribution, expanding Beta random variable to multi-dimension. It is also a special case of Dirichlet process when the dimension is finite and fixed.
\begin{defn}
A $K$-dimensional continuous random vector $(\pi_1,...,\pi_K)$ follows Dirichlet distribution with parameters $(\phi_1,...,\phi_K)$, if $\pi_k\in[0,1]$ for all $k$ and $\sum_{k=1}^{K}\pi_k=1$; its probability density function can be expressed as
\[
\begin{aligned}
    &(\pi_1,...,\pi_K)\sim\textup{Dirichlet}(\phi_1,...,\phi_K)\\ &p(\pi_1,...,\pi_K)=\frac{\Gamma(\sum_{k=1}^{K}\phi_k)}{\prod_{k=1}^{K}\Gamma(\phi_k)}\prod_{k=1}^{K}\pi_k^{\phi_k-1}
\end{aligned}
\]
\end{defn}
$\phi_k>0$ for all $k$. Dirichlet distribution is the conjugate prior of Multinomial distribution.
\end{appendices}

\end{document}